\documentclass[review]{elsarticle}

\usepackage{lineno,hyperref}
\usepackage{multirow}

\usepackage{times}
\usepackage{latexsym}
\usepackage{graphicx}
\usepackage{caption}
\usepackage{subcaption}
\usepackage{algorithm}
\usepackage{algorithmic}
\usepackage{amsmath}

\graphicspath{{./imgs/}}
\usepackage{booktabs}
\usepackage{tabularx}

\newcommand\clearrow{\global\let\rowmac\relax}
\clearrow

\journal{Journal of Cognitive Systems Research}

\begin{document}

\begin{frontmatter}

\title{Hierarchical growing grid networks for skeleton based action recognition}

\author{Zahra Gharaee\corref{mycorrespondingauthor}}
\address{Cognitive Science, Department of Philosophy \\ University of Lund, Lund, Sweden}\fnref{mycorrespondingauthor}
\address{Computer Vision Labarotory (CVL), Department of Electrical
	Engineering \\ University of Link\"{o}ping, Link\"{o}ping, Sweden }\fnref{mycorrespondingauthor}
\fntext[myfootnote]{Computer Vision Labarotory (CVL)\\ Link\"{o}ping University
	58183 Link\"{o}ping, Sweden}


\cortext[mycorrespondingauthor]{Corresponding author}
\ead{zahra.gharaee@liu.se}


\begin{abstract}
In this paper, a novel architecture for action recognition is developed by applying layers of growing grid neural networks. Using these layers makes the system capable of automatically arranging its representational structure. In addition to the expansion of the neural map during the growth phase, the system is provided with a prior knowledge of the input space, which increases the processing speed of the learning phase. Apart from two layers of growing grid networks the architecture is composed of a preprocessing layer, an ordered vector representation layer and a one-layer supervised neural network. These layers are designed to solve the action recognition problem. The first-layer growing grid receives the input data of human actions and the neural map generates an action pattern vector representing each action sequence by connecting the elicited activation of the trained map. The pattern vectors are then sent to the ordered vector representation layer to build the time-invariant input vectors of key activations for the second-layer growing grid. The second-layer growing grid categorizes the input vectors to the corresponding action clusters/sub-clusters and finally the one-layer supervised neural network labels the shaped clusters with action labels. Three experiments using different data sets of actions show that the system is capable of learning to categorize the actions quickly and efficiently. The performance of the growing grid architecture is compared with the results from a system based on Self-Organizing Maps, showing that the growing grid architecture performs significantly superior on the action recognition tasks.
\end{abstract}


\end{frontmatter}


\section{Introduction}
\label{intro}
Action recognition is important in our daily lives since it is necessary for understanding the behavior of others. We perceive an action by observing the kinematics of the body parts involved in the performance. We use our experience and concepts to make a correct categorization of the action. Although learning the action concepts is a life-long process, we behave very efficiently in applying our learned concepts in analyzing motions and recognizing actions.  

The experiments performed by using the patch light technique designed by Johansson \cite{johansson} show that an action is recognized after only about two hundred milliseconds of the observation. More detailed features of the action performed, such as the gender of the performer or the weight of the lifted object (where the objects were not visible), were perceived by the observer by just watching the films of the moving dots representing some skeleton joints of the action performer (see \cite{Runesson1} and \cite{Runesson2}).

Since action recognition is so efficient in humans, it is a challenge to construct an artificial system that can perform a similar task. This is an important challenge, since there are numerous applications for an action recognition system such as video surveillance, human-computer interaction, sign language recognition, robotics, video analysis (sports video analysis) and entertainment industry. Therefore a robust action recognition system must be validated for different sources of input method containing different types of actions. 

The evaluation of the performance of an artificial action recognition system depends on several factors. Two common measure that are used to validate a system performance are the processing speed for training the system and the accuracy of the trained system when generalizing the learned concepts in test experiments involving different kinds of actions. The learning speed is calculated through the time it takes the system model to regulate its parameters. 

Another important factor when choosing the approach to an action recognition system is how biologically plausible it is since the action recognition task is performed excellently by the biological systems. In particular, \cite{johansson} patch-light technique for analyzing biological motion shows that through watching the films - in which only the moving dots of light could be seen - subjects recognized the actions within tenths of a second. An important lesson to learn from the experiments by Johansson and his followers (\cite{Runesson1} and \cite{Runesson2}) is that the kinematics of a movement contains sufficient information to identify the underlying dynamic patterns. Thus a system that resembles the neuronal system in animals or humans may give us a better understanding of how organisms categorize actions. A possible goal of an artificial system can be to perform in an optimal and efficient way similar to the performance of living organism such as humans or animals.

In this article I propose a novel multi-layer architecture for human action recognition, which is composed of several processing layers including two layers of growing grid neural networks. As a background to the artificial neural networks utilized in this study, I first present the cortical mechanisms representing the peripheral input space. Next, I will present the simplified but useful computational models such as self-organizing maps and growing grid networks that to some extent mimic the architecture of the brain. Later in section 2 the proposed architecture is described and the experiments on action recognition are presented in detail in section 3. A comparison with other action recognition systems is also provided in section 4. Finally the section 5 concludes the paper.

\subsection{Cortical mechanisms representing the peripheral input space}
As we argued above, observations from biological systems are highly relevant when designing technical solutions. We know that cortical representations in adult animals are not static and fixed entities but dynamic and modified throughout life. The plastic changes occurring at synaptic level (cortical synaptic plasticity or Hebbian plasticity \cite{Hebb}) involve an increase in synaptic strength between neurons that fire together. A higher level of plasticity is the reorganization of cortical representations, based on the Hebbian-based learning rules, in which the temporally correlated inputs are detected. This entails that the inputs from peripheral resources that fire in close temporal proximity are more likely to represent neighboring areas in the sensory cortex \cite{Buonomano}.

In addition to the vertical flow of information connecting the peripheral sensory input to the cortex, there exists a horizontal inter-connectivity that integrates the information from neighboring regions and from specific to more distal cortical zones \cite{Buonomano}. The reorganization of cortical maps can be related to such a horizontal connectivity between neighboring cortical sectors.  

In general the growth of a network relates to an increase in any types of its components. Since in the nervous system the neurons and the synapses are among the network components, an increase in the number of neurons or the number and/or strength of the synapses will lead to the growth of the nervous system. Although increasing complexity is a natural consequence of a growing network, it also increases the abilities of the network to resolve more difficult and complicated problems, which makes the growth necessary and desirable. 

The cortical reorganizations occur as a result of peripheral or central alterations of inputs and in response to behavior. The cortex can dynamically allocate an area in a use-dependent manner to inputs that have different levels of engagement. As an example, one study shows an almost twofold expansion of the cortical representation of nipple-bearing skin in lactating female rats compared with non-lactating female rats \cite{Buonomano}. 

The results of digit amputation in adult monkeys presented in \cite{Merzenich} also show that around two to eight months after the amputation, most of the cortex area that responds to the amputated digit(s) in control animals now respond to the adjacent digits or the subjacent palm in the amputated animals. This shows an expansion of cortical representation for parts of the input space that are mostly used (non-amputated areas adjacent to the amputated ones).

\subsection{Self-organising maps}
A useful model that has properties that to some extent explain for instance the topographical mapping is the Kohonen feature map \cite{kohonen} or self-organizing map (SOM). Important properties are layered and topographic organization of the neurons, lateral interactions, Hebb-like synaptic plasticity, and the capability of unsupervised learning. A Kohonen feature map is a two-dimensional square grid of neurons with a fixed number and a fixed topology. All neurons of the map receive input in parallel from the sensory input space.

The links connecting the neurons represent the horizontal inter-connectivity between them which is modeled by a neighboring function, for example a Gaussian function. There is no exchange of data through these links and they are just used to represent the neighborhood relationship. The data exchange occurs in parallel only between the receptors and the neurons while each neuron of the map has a connection to each receptor, which resembles the preferred vertical flow of information in the human sensory cortex.  

For each input signal the winner is the neuron in the whole map with the nearest reference vector. Then its reference vector together with its topological neighbors’ reference vectors are updated in such a way that they are moved towards the input signal. After several adaptation steps, similar input signals will be mapped onto the neighboring areas of the network – known as topographic mapping. In the same way, in the human nervous system the sensory cortical areas of touch, vision and hearing represent their receptive sensory epithelial surfaces in a topographical manner \cite{Gazzaniga}. Thus, adjacent areas of the peripheral sensory space such as adjacent fingers are mapped onto the neighboring regions of the sensory cortex.

One of the main features of the self-organizing neural networks is their ability to generate low-dimensional representations of high-dimensional input spaces. This feature is important in applications in which there is a high dimensional input to the system such as in image processing. A wide range of applications utilizing the self organizing maps \cite{kohonen2} have been developed, such as vector quantization and image compression \cite{Dony}, biological modeling and parallel computing \cite{Obermayer}, and combinatorial solutions for an optimization problem \cite{Favata}. 
The proposed method in \cite{parisi2016human} also uses self-organizing networks to assess the quality of actions performed. Their learning-based method provides feedback on a set of training movements, 3 powerlifting exercises performed by 17 athletes, captured by a depth sensor.

\subsection{Growing grid networks}
Despite all the advantages of the Kohonen feature map, the implementation of the architecture presupposes a predetermined size in terms of the number of rows and columns of the network. This results in less flexibility in the self-organizing feature maps.

The growing hierarchical self-organizing map (GHSOM) proposed by \cite{Dittenbach1} is an approach to address two limitations of the SOM systems: the static architecture of the SOM model as well as its limitations to represent hierarchical relations of the input. To the best of our knowledge, GHSOM has not been tested on the spatio-temporal input space and it has been used to classify documents, which is represented in \cite{Dittenbach2}.

On the other hand in order to generate a precise representation of the topology, a priori knowledge of the input space is required. Building this knowledge demands a proportionally high computational effort, especially in more realistic experiments. Therefore, if an algorithm exploits some effective heuristics through the peripheral sensory input in order to guide the development of the architecture, then a more accurate topological representation of the input space is reachable \cite{Blackmore}. 

The growing grid network structure proposed by Fritzke meets these requirements \cite{Fritzke1}, \cite{Fritzke5}. Such growing networks been applied to a classification problem \cite{Fritzke2}, to a combinatorial optimization problem \cite{Fritzke3}, to a problem of surface reconstruction \cite{Ivrissimtzis} and also to the touch perception in a robotic task \cite{Johnsson4}. 

A comparison of self-organizing maps and the growing grid structures is presented in \cite{Fritzke4} and the results show that, although the self-organizing feature maps achieve a slightly better performance in the simplest tasks, the growing grid structures perform significantly better in more complicated problems, which is the case of more realistic action recognition experiments.

The main contributions of this article are listed as:\\
(1) A novel cognitive architecture for 3D skeleton based human action recognition based on the growing grid neural networks is proposed.\\
(2) The proposed architecture is evaluated on three different public datasets and it efficiently reaches quite high performance in recognizing human actions.\\
(3) The proposed architecture is compared with an action recognition architecture based on the self-organizing maps in terms of training time and accuracy. The results approve that using growing grid networks, the action recognition task is performed more efficiently. This can be because the growing grid layers make the system capable of automatically arranging its representational structure. Moreover, the system is provided with a prior knowledge of the input space, which increases the processing speed of the learning phase.

\section{Related works on skeleton based human action recognition}

Using the skeleton based data, the cost-effective depth sensors are coupled with the real-time 3D skeleton estimation algorithm. Most of the skeleton-based methods utilize either the 3D locations or the angles of the joints to represent the human skeleton. One can find quite a number of research studies on skeleton based human action recognition. To start with, I refer to the earlier works by the author \cite{Gharaee2, Gharaee3, Gharaee4, Gharaee5, Gharaee6, Gharaee7}, which address various aspects of action recognition problem.

By extracting the spatial-temporal features from the 3D skeleton information, such as the relative geometric velocity between body parts, relative joint positions and joint angles in \cite{Yao}, the position differences of the skeleton joints in \cite{Yang-Xiaodong1} or the pose information together with differential quantities (speed and acceleration) in \cite{Zanfir}, the body skeleton information in space and time is first described. Then the descriptors are coupled with Principle Component Analysis (PCA) or some other classifier to categorize the actions. There are other methods in the literature using skeleton data for human action recognition such as \cite{Chaudhry, Wang-Chunyu, vemulapalli2014human}.

The Growing When Required (GWR) networks proposed by \cite{parisi2015self} consists of a two-stream hierarchy of self-organizing growing networks, which processes pose and motion features in parallel and subsequently integrates clustered neuronal activation trajectories from both streams. The GWR starts with a set of two nodes randomly initialized and at each time step, both
the nodes and the edges can be created and removed. 

In some other methods, a fusion-based feature for the action recognition is applied, for example the method proposed by \cite{Zhu2013fusing} in which the spatio-temporal features and the skeleton joints are fused as complementary features to recognize human actions. Another method that uses multi-fused features to recognize human actions is the Human Activity Recognition (HAR) system proposed by \cite{Jalal}. This method fuses four skeleton joint features together with one body shape feature representing the projections of the depth differential silhouettes between two consecutive frames onto three orthogonal planes.

In recent works by \cite{shi2019skeleton, shi2019two} a graph based neural networks approach is proposed for human action recognition. In \cite{shi2019skeleton}, the authors propose the skeleton data as a directed acyclic graph (DAG) based on the kinematic dependency between
the joints and bones in the natural human posture. They use a neural network based graph to extract the information of joints, bones and their relationships and to predict the extracted features. In \cite{shi2019skeleton}, two-stream adaptive graph convolutional network for skeleton-based action recognition is used. The topology of the graph is learned either uniformly or individually in an end-to-end manner to increase the flexibility of the model for graph construction and to adapt to various data samples.  In \cite{si2019attention} also, an Attention Enhanced Graph Convolutional LSTM Network is proposed for human action recognition using skeleton data. The proposed approach captures discriminative features in spatial configuration and temporal dynamics and it explores the co-occurrence relationship between spatial and temporal domains.

 Finally, the method by \cite{zhang2019view} designs two view adaptive neural networks, which are respectively built based on the recurrent neural networks with the Long Short-term Memory (LSTM) and the convolutional neural network. For each network, a novel view adaptation module learns to determine the best observation viewpoints, and transforms the skeletons to those viewpoints for the end-to-end recognition with a main classification network. 

and \cite{zhang2019view}

\begin{figure*}
	\centering
	\includegraphics[width=1.00\textwidth]{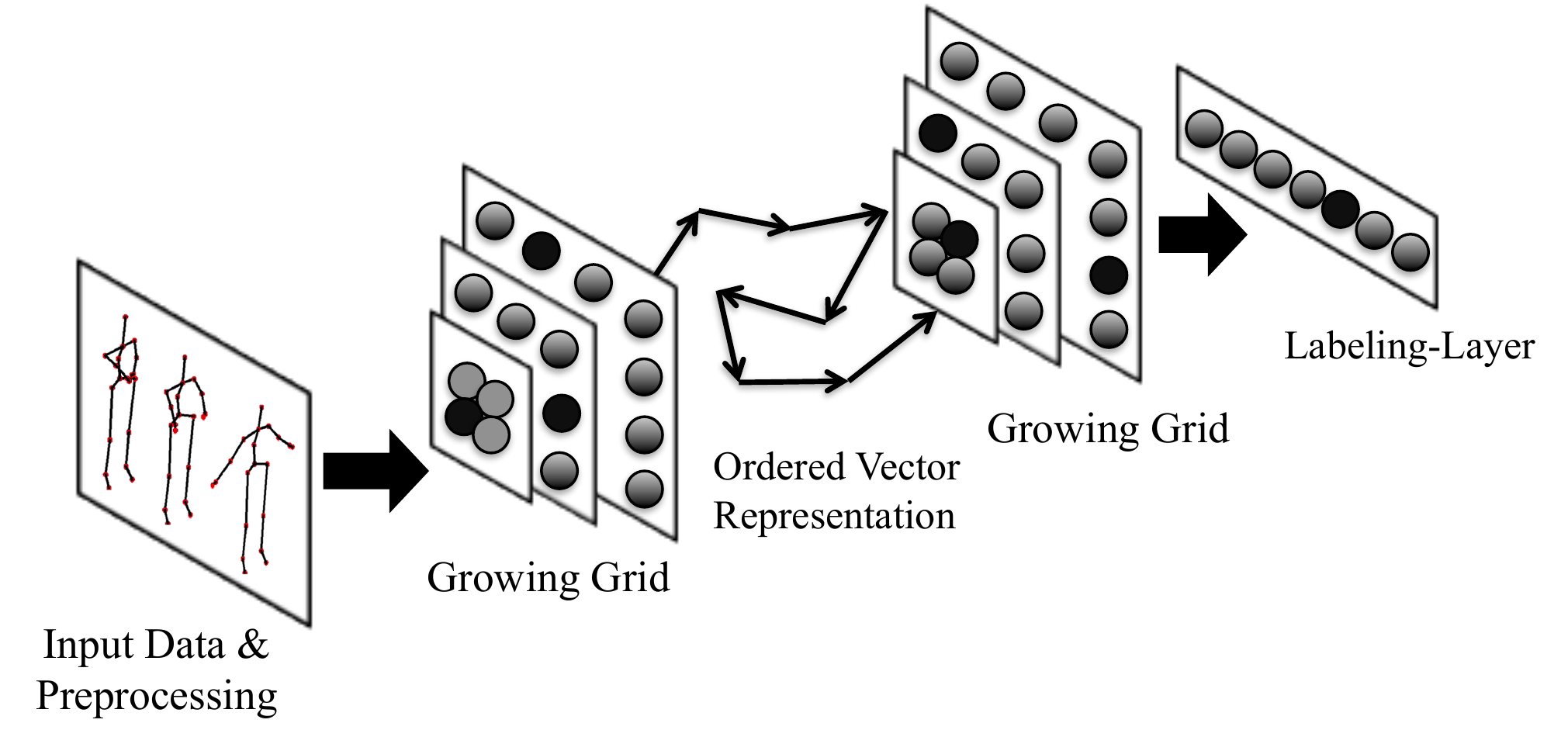}
	\caption{The hierarchical growing grid architecture for recognizing and clustering human actions. The architecture is composed of five processing layers including three layers of neural networks. The first and second neural network layers consist of growing grid and the third one is a one-layer supervised neural network to label the action categories made by the second-layer growing grid.  }
	\label{fig:HGG}       
\end{figure*}

\section{Architecture}
\label{sec:3}

In this section the hierarchical architecture shown in Fig.~\ref{fig:HGG} is described. The architecture consists of three layers of neural networks, in addition to a layer of preprocessing and a layer of ordered vector representation. The ordered vector representation layer is utilized to build time-invariant action pattern vectors. As a result of this implementation, the corresponding patterns achieved from the first-layer growing grid are invariant to the speed of performing different actions.

\subsection{Input data and preprocessing}

The input data of actions for the experiments of this paper are generated by RGB-D sensors. The recent development of such sensors (for example, $Microsoft Kinect^{TM}$ and $Asus Xtion^{TM}$) lead to motion recognition systems that attract much more attention due to the extra dimension provided by depth, which is less sensitive to the illumination and color changes and also includes 3D information of the scene.

The dense neighborhood in RGB data contain information about color and texture. Moreover, it enables the extraction of interest points and optical flow. The depth data is insensitive to the illumination changes, invariant to color and changes of texture, and provides us with 3D information. There are neural network based approaches using either the RGB data as the input space, such as \cite{Pigou} and \cite{Zolfaghari} or the depth data proposed by \cite{Wang-Pichao} and \cite{Rahmani}. 

The third input type, used in this article is the skeleton data containing the positions of the human joints, which are relatively high-level features for motion recognition. Moreover, the skeletal data is more robust to scale, illumination, and color changes and can be made invariant of camera view as well as the rotation of the body. There exist systems using skeleton information as the input data for the action recognition problem such as the Convolutional Neural Network (CNN-based) approaches proposed by \cite{Liu} and \cite{Hou}, the Recurrent Neural Network (RNN-based) approaches proposed by \cite{Du} and \cite{Veeriah}, and other types of neural network based systems such as the one proposed by \cite{Ijjina}.

The methods used to extract the input data of the actions performed such as the 3D information of the skeleton joints deals with the action detection problem, that is, the problem of detecting the moving figures. This problem must be solved before the action recognition can be initiated. However, the action detection problem is not addressed in this study. 

\subsubsection{Attention} The preprocessing layer, which receives the input data from the action detection module executes several functions. Among them is an attention mechanism that is inspired by human behavior, paying attention to the most salient parts of the body when recognizing an action \cite{Gharaee1}. The saliency in this work is determined by the movement (velocity). The skeleton posture of the performer is divided into five main parts: left arm, left leg, right arm, right leg and the base (including head, neck, torso and stomach). The body part with the largest movement during acting receives the most attention (attention focus) and the rest of body is ignored. This means that the system receives and processes only the postural information of where the attention is focud.

\subsubsection{Ego-Centered Coordinate Transformation}Preprocessing module also includes an ego-centered coordinate transformation to make input data invariant of having different orientations toward the camera. The new coordinate system is called an ego-centered coordinate system because its origin is located in the joint Stomach of the performer. The three joints Stomach, Left Hip and Right Hip are utilized to build the axis of the new right handed coordinate system as shown in see Fig.~\ref{fig:ego}.

\begin{figure}
	\centering
	\begin{subfigure}[t]{0.15\linewidth}
		\includegraphics[width=\textwidth]{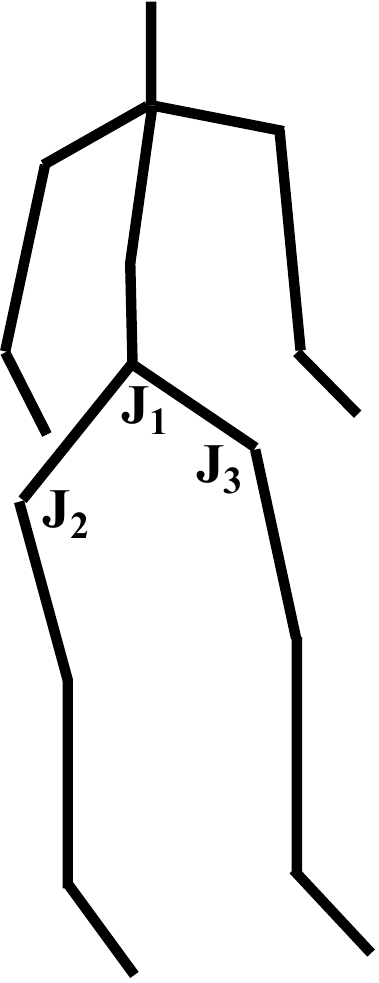}
		\caption{}
		\label{fig:pos}
	\end{subfigure}
	\begin{subfigure}[t]{0.56\linewidth}
		\includegraphics[width=\textwidth]{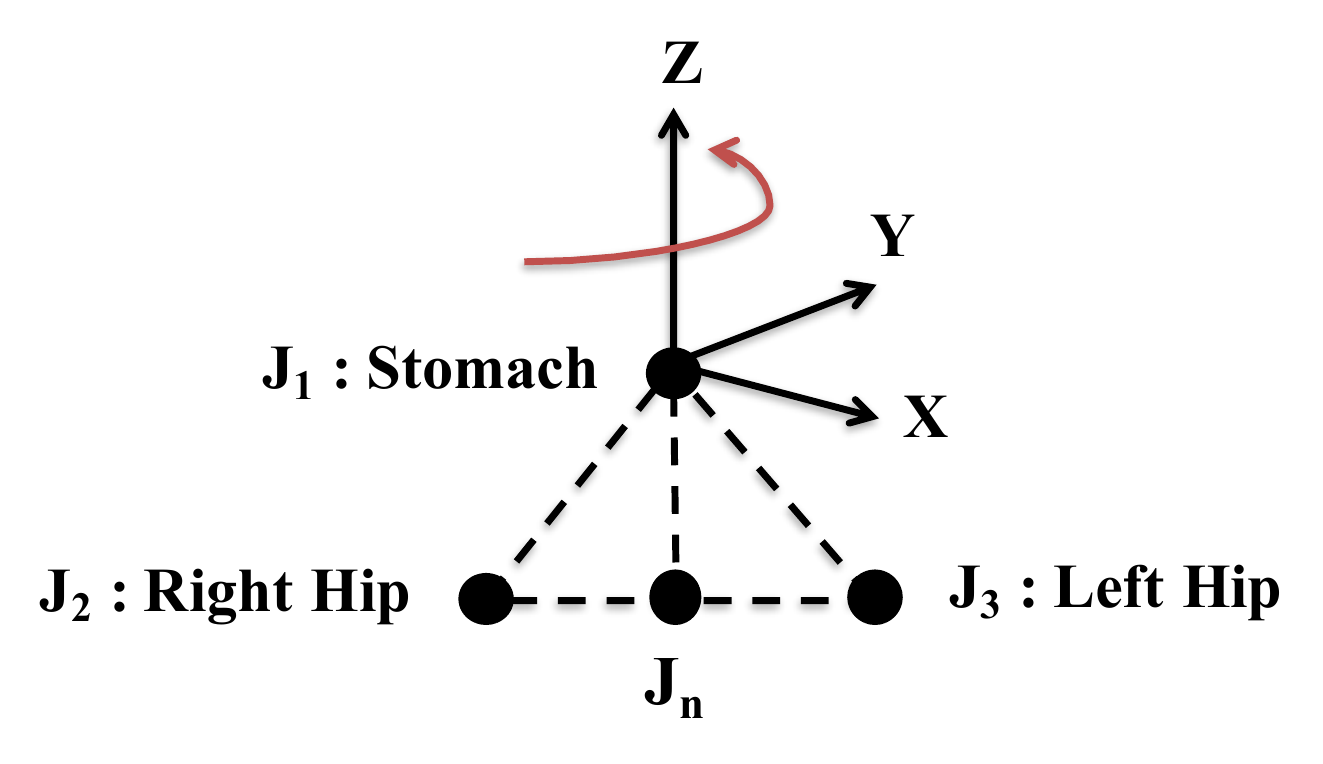}
		\caption{}	
		\label{fig:ego}
	\end{subfigure}
	\caption{The body posture consists of 3D skeleton joints information (a). The new coordinate system for the body posture, an ego-centered coordinate system located at the joint Stomach of the performer, built from the joints Stomach, Right Hip and Left Hip (b).}
	\label{fig:ego}
\end{figure}

To transform to the ego-centered coordinate system, first the projection of the joint Stomach $J_1$ on the line connecting joints right hip $J_2$ and left hip $J_3$ is calculated and called $J_n$. Precise location of $J_n$ is calculated by solving the system equations defined as follows if we assume $J_1 = (x_1, y_1, z_1)$, $J_2 = (x_2, y_2, z_2)$ and $J_n = (x_n, y_n, z_n)$:

\begin{equation}\label{eq:1}
sys = \begin{cases}
Eq_1: \overrightarrow{N} = [n_x, n_y, n_z] = \frac{\mathbf (\overrightarrow{J_3-J_2})}{\left\Vert\left( \mathbf (\overrightarrow{J_3-J_2}) \rm\right) \right\Vert}\\
Eq_2: \begin{cases}
x_n = n_x \times t + x_2 \\
y_n = n_y \times t + y_2\\
z_n = n_z \times t + z_2\\
\end{cases} \\
Eq_3: \mathbf (\overrightarrow{J_n-J_1}) \cdot \mathbf (\overrightarrow{J_3-J_2}) = 0 \\
\end{cases}
\end{equation}

Having (\ref{eq:1}), $Eq_1$ gives the normal vector of the line connecting $J_2$ and $J_3$ and $Eq_2$ defines $J_n$ 3D coordinates as standard line equation and therefore it remains only one unknown parameter $t$ to be calculated. Finally, the unknown parameter $t$ is given by solving $Eq_3$, which is the dot product of the vectors $\overrightarrow{J_n-J_1}$ and $\overrightarrow{J_3-J_2}$. Since the projection is the closest point, these vectors are orthogonal. 

When $J_n$ coordinates are calculated, unit vectors of the ego-centered coordinate system $\overrightarrow{x_E}$, $\overrightarrow{y_E}$ and $\overrightarrow{z_E}$ representing its 3D-axis are given by:

\begin{equation}\label{eq:2}
 U_{E} = \begin{cases}
\overrightarrow{x_{E}} = \frac{\mathbf (\overrightarrow{J_3-J_n}) \times \mathbf (\overrightarrow{J_1-J_n})}{ \left\Vert\left( \mathbf (\overrightarrow{J_3-J_n}) \times \mathbf (\overrightarrow{J_1-J_n})  \rm\right) \right\Vert}\\
\overrightarrow{y_{E}} = \frac{\mathbf (\overrightarrow{J_3-J_n})}{ \left\Vert\left( \mathbf (\overrightarrow{J_3-J_n}) \rm\right) \right\Vert } \\
\overrightarrow{z_{E}} = \frac{\mathbf (\overrightarrow{J_1-J_n})}{\left\Vert\left( \mathbf (\overrightarrow{J_1-J_n}) \rm\right) \right\Vert}
\end{cases}
\end{equation}

Having (\ref{eq:2}), $\overrightarrow{x_E}$ is the cross/vector product of the vectors $\mathbf (\overrightarrow{J_3-J_n})$ and $\mathbf (\overrightarrow{J_1-J_n})$. The unit vectors of the Reference coordinate system are known as:

\begin{equation}\label{eq:3}
U_{R} = \begin{cases}
\overrightarrow{X_{R}} = [1,  0,  0]\\
\overrightarrow{Y_{R}} = [0,  1,  0] \\
\overrightarrow{Z_{R}} =[0,  0,  1]
\end{cases}
\end{equation}

In order to get a 3D joint in the ego-centered coordinate system ${}^{E}J$, the rotation matrix ${}^{R}_{E}R $ is calculated by dot product of the pair of unit vectors as its components:

\begin{equation}\label{eq:4}
{}^{R}_{E}R = 
\begin{pmatrix}
{}^{R}\overrightarrow{X_E} & {}^{R}\overrightarrow{Y_E} & {}^{R}\overrightarrow{Z_E}
\end{pmatrix}
=
\begin{pmatrix}
\overrightarrow{x_E} \cdot \overrightarrow{X_R} & \overrightarrow{y_E} \cdot \overrightarrow{X_R} & \overrightarrow{z_E} \cdot \overrightarrow{X_R} \\
\overrightarrow{x_E} \cdot \overrightarrow{Y_R} & \overrightarrow{y_E} \cdot \overrightarrow{Y_R} & \overrightarrow{z_E} \cdot \overrightarrow{Y_R} \\
\overrightarrow{x_E} \cdot \overrightarrow{Z_R} & \overrightarrow{y_E} \cdot \overrightarrow{Z_R} & \overrightarrow{z_E} \cdot \overrightarrow{Z_R} 
\end{pmatrix}.
\end{equation}

The transformation of the joint coordinates to the ego-centered frame ${}^{E}J$ is then calculated as follows:

\begin{equation}\label{eq:5}
\begin{pmatrix}
{}^{R}J  \\
1
\end{pmatrix}
= 
\begin{pmatrix}
	\begin{array}{c|c}
	{}^{R}_{E}R & {}^{R}J_{org} \\ 
	\hline
	0 0 0  & 1
	\end{array}
\end{pmatrix}
\begin{pmatrix}
{}^{E}J  \\
1
\end{pmatrix},
\end{equation}

where ${}^{R}J$ is a joint coordinates in the Reference frame and ${}^{E}J$ is its transformation to the ego-centered frame. Term ${}^{R}J_{org}$ is the origin of the ego-centered frame, which is replaced by the joint stomach ($J_1$ in Fig. \ref{fig:ego}). All joints 3D coordinates are transformed into this ego-centered coordinate system using  (\ref{eq:5}).

\subsubsection{Scaling} The scaling function is designed to make the input data invariant of having different distances to the camera. All posture 3D information is scaled into one posture size. Assume each body posture is composed of 20 joints and 19 links connecting each pair of joints. There is a fixed length defined for each link and all posture frames are scaled to have the pre-defined values for their links. 

Assume a consecutive pair of joints like stomach $J_1$ and right hip $J_2$ shown in Fig. \ref{fig:ego} having a fixed length $L$ of the connecting link. In order to scale the original link to have the length $L$, the position of a new joint coordinates $J_n = [x_n, y_n, z_n]$ is calculated to replace the joint right hip $J_2$. The coordinates of $J_n$ is achieved by solving the system equations as:

\begin{equation}\label{eq:6}
sys = \begin{cases}
Eq_1: N = [n_x, n_y, n_z] = \frac{\mathbf (\overrightarrow{J_2-J_1})}{\left\Vert\left( \mathbf (\overrightarrow{J_2-J_1}) \rm\right) \right\Vert}\\
Eq_2: \begin{cases}
x_n = n_x \times t + x_1 \\
y_n = n_y \times t + y_1\\
z_n = n_z \times t + z_1\\
\end{cases} \\
Eq_3: \left\Vert\left( J_{n}-J_{1} \rm\right) \right\Vert = L\\
\end{cases}
\end{equation}

Having (\ref{eq:6}), $Eq_1$ is the normal vector of the line connecting $J_1$ and $J_2$. Since $J_n$ must be located on this line, it satisfies the standard line equation from $Eq_2$. As a result, it remains only one unknown parameter $t$ to get 3D coordinates of $J_n$. Parameter $t$ is given by solving $Eq_3$, which determines the distance criteria to re-scale the link in order to have the length $L$.

\subsection{Growing grid mechanisms}
\label{sec:PreP}
A growing grid network is an incremental variant of self-organizing feature maps that contains an increasing number of neurons with a fixed topology. In the growing grid, the network grows by insertion of new rows or columns at certain time intervals during learning. The new row or column will be inserted in the locations of the network where the input space has more complexity and the system requires a larger area to represent the input space. 

The growing grid learns to represent the input space in two phases: the growth phase and the fine tuning phase (see Fig.~\ref{fig:GG}). During the growth phase, the rectangular network begins with a minimum number of neurons ($2\times 2$) and, by inserting a complete row or column, the network size increases until a performance criterion is met (for example, a maximum number of neurons).

The fine tuning phase starts immediately after the network meets the performance criterion. The size of the network achieved at the end of the growth phase does not change any more during this phase. The fine-tuning phase continues learning with a fixed number of rows and columns and a decaying learning rate to find the good final values of the input data.

\subsubsection{Growth Phase}
\label{secgg}
\begin{figure*}
	\centering
	\includegraphics[width=0.8\textwidth]{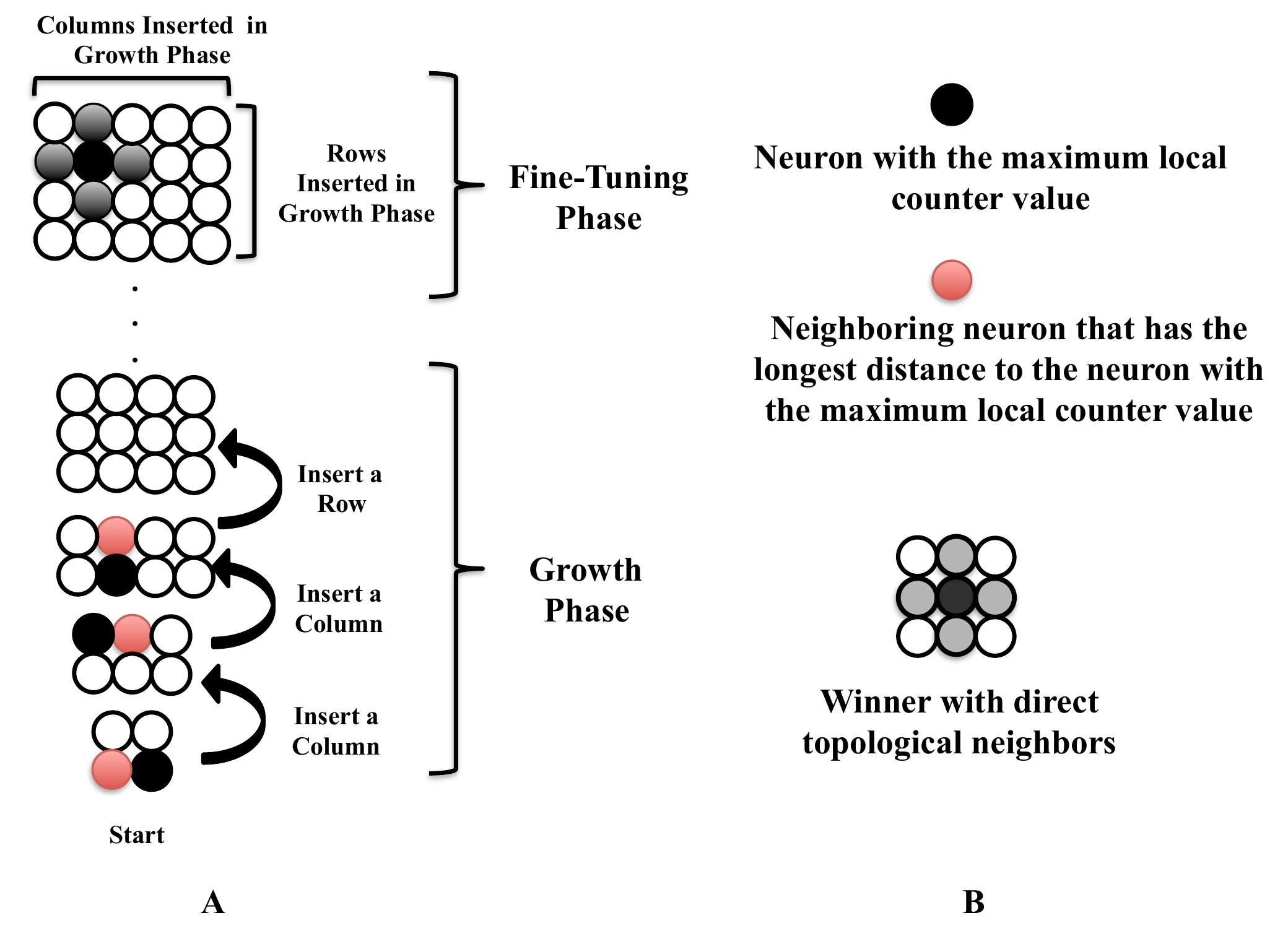}
	\caption{Part A of the figure shows how the growing grid implementation starts from a rectangular grid of $2\times 2$ neurons and then inserts a new row or column in each adaptation step during the growth phase. The growth phase continues learning until a performance criterion is met and then the fine-tuning phase starts to learn the input space and regulate the network parameters with decaying learning rate and a fixed topology. Part B: The top first row represents the neuron that has been activated the most during an adaptation step, the second row shows the neuron with the largest distance to the most activated neuron, which is detected among direct topological neighbors and the last row shows the direct topological neighborhood (shown in gray) of a winner neuron (shown in black).}
	\label{fig:GG}       
\end{figure*}

The growth phase starts with a network of rectangular shape as shown in Fig.~\ref{fig:GG} (part A), with the size of $2\times 2$ in which each neuron $n_{ij}$ is associated with a weight vector $w_{ij}\in{R}^n$ with the same dimensionality as the input vectors. All the elements of the initial weight vectors are initialized by real numbers randomly selected from a uniform distribution between 0 and 1. In addition to a weight vector $w_{ij}\in{R}^n$ each neuron has a local counter variable $LC_{ij}$ to estimate the location of a new insertion of a row or column. 

At time $t$ the input vector $x(t)\in{R}^n$ is received by each neuron of the network representing the parallel computations in a growing grid network. The neuron $w_{c}$ that is the most similar to the input vector $x(t)$ is selected by:  

\begin{equation}\label{eq:7}
w_{c}=\mathrm {arg} \mathrm{ max}_{ij}y_{ij}(t),
\end{equation}

where $y_{ij}=e^{\frac{-s_{ij}(t)}{\sigma}}$ refers to the activity of each neuron calculated by applying the exponential function to the net input $s_{ij}$ and $\sigma$ is the exponential factor used to normalize and increase the contrast between highly activated and less activated areas. The value of $\sigma$ depends on the input data, for example the $\sigma$ is set to ${10^6}$ in the first layer growing grid. Net input $s_{ij}$ is calculated by applying the Euclidean metric to each input vector and the weight vector of each neuron:

\begin{equation}\label{eq:9}
s_{ij}(t)=||x(t) - w_{ij}(t)||,
\end{equation}

where $i$ and $j$ represent the corresponding row and column of a neuron and are $0 \leq {i} < I$, $0 \leq {j} < J$, ${i},{j}\in{N}$. After finding the neuron $w_{c}$, its local counter variable is incremented by one ($LC_{w_{c}}=LC_{w_{c}}+1$) and the weight vectors $w_{ij}$ associated with $w_{c}$ and the neurons $n_{ij}$ in its direct topological neighborhood as shown in Fig.~\ref{fig:GG} (part B) are updated:

\begin{equation}\label{eq:11}
w_{ij}(t+1)=w_{ij}(t)+\alpha[x(t)-w_{ij}(t)]
\end{equation}

The learning rate $\alpha$ is a constant and is not a function of time for the growth phase. Parameters $p$ and $q$ determine the locations of the neurons in the direct topological neighborhood of $w_{c}$. If the row and column of $w_{c}$ are defined by $p_{c}$ and $q_{c}$ then the set of the direct topological neighbors of $w_{c}$ equals:

\begin{equation}\label{eq:115}
DTN:\big\{n_{p_{c}-1,q_{c}},n_{p_{c}+1,q_{c}},n_{p_{c},q_{c}-1},n_{p_{c},q_{c}+1}\big\}.
\end{equation}

A major component of the growing grid networks is to insert new neurons. The value of $\lambda$ determines the time for a new insertion. If the $\lambda$ is too small the net grows too fast before it has adapted enough to the input space and if it is too large then the net grows slowly due to the lack of insertion. For the experiments of this study the \textit{middle approach} is used to set the lambda, which means that the lambda is set in the middle of the total length of the input space. In other words, the new insertion occurs when half of the input data is met by the network. Therefore there will be maximum 2 insertions per epoch. When the $\lambda$ criterion is met the neuron $w_{c_{1}}$ with the largest local counter value $LC$ is given as:

\begin{equation}\label{eq:12}
w_{c_{1}}=\mathrm {arg} \mathrm{ max}_{ij}LC_{ij},
\end{equation}

and, among its direct topological neighbors $DTN$, the neuron $w_{c_{2}}$ with the furthest distance to the $w_{c_{1}}$ is selected by:

\begin{equation}\label{eq:13}
w_{c_{2}}(t)=\mathrm {arg} \mathrm{ max}_{w_{n}\subset DTN}||w_{w_{c_{1}}}(t)-w_{n}(t)||
\end{equation}

  If both neurons $w_{c_{1}}$ and $w_{c_{2}}$ are in the same row then a new column is inserted between them and the weight vectors of the neurons of the new column are an interpolation of the weight values of the neurons in the neighboring columns. Similarly, when the neurons $w_{c_{1}}$ and $w_{c_{2}}$ are in the same column, then a new row is inserted between them and the weight vectors of the neurons in the new row are an interpolation of the weight values of the neurons in the neighboring rows.

When the insertion is completed, the local counter value $LC_{ij}$ and the $\lambda$ are reset. The growth phase continues until a performance criterion is met by $\gamma$.
Detailed description of how $\lambda$ and $\gamma$ criterion are set for the aims of this article is available under the section \ref{paramset}.

\subsubsection{Fine-tuning phase}
In the fine-tuning phase, the same principles are applied as in the growth phase. The adaptation strength (learning rate ) is a function of time $\alpha(t)$, which is decaying so the updates are done as:

\begin{equation}\label{eq:14}
w_{ij}(t+1)=w_{ij}(t)+\alpha(t)[x(t)-w_{ij}(t)],
\end{equation}

where $i$ and $j$ represent the corresponding row and column of a neuron and are $0 \leq {i} < I$, $0 \leq {j} < J$, ${i},{j}\in{N}$. Moreover, there is no insertion of new neurons. The net size represented by the number of rows and columns is maintained from the growth phase.
The network continues learning in the fine-tuning phase for a number of steps with a fixed size and topology to regulate all its parameters based on the input data.

\subsection{Ordered vector representation}
To communicate the first and the second layer growing grid an ordered vector representation module is designed and implemented, which creates time invariant pattern vectors from activity traces of the first-layer growing grid. Each input vector activates one neuron of the first-layer growing grid network and as a result the consecutive input vectors representing posture frames of an action sequence creates an activity pattern vector.

\begin{figure*}
	\centering
	\includegraphics[width=1.0\textwidth]{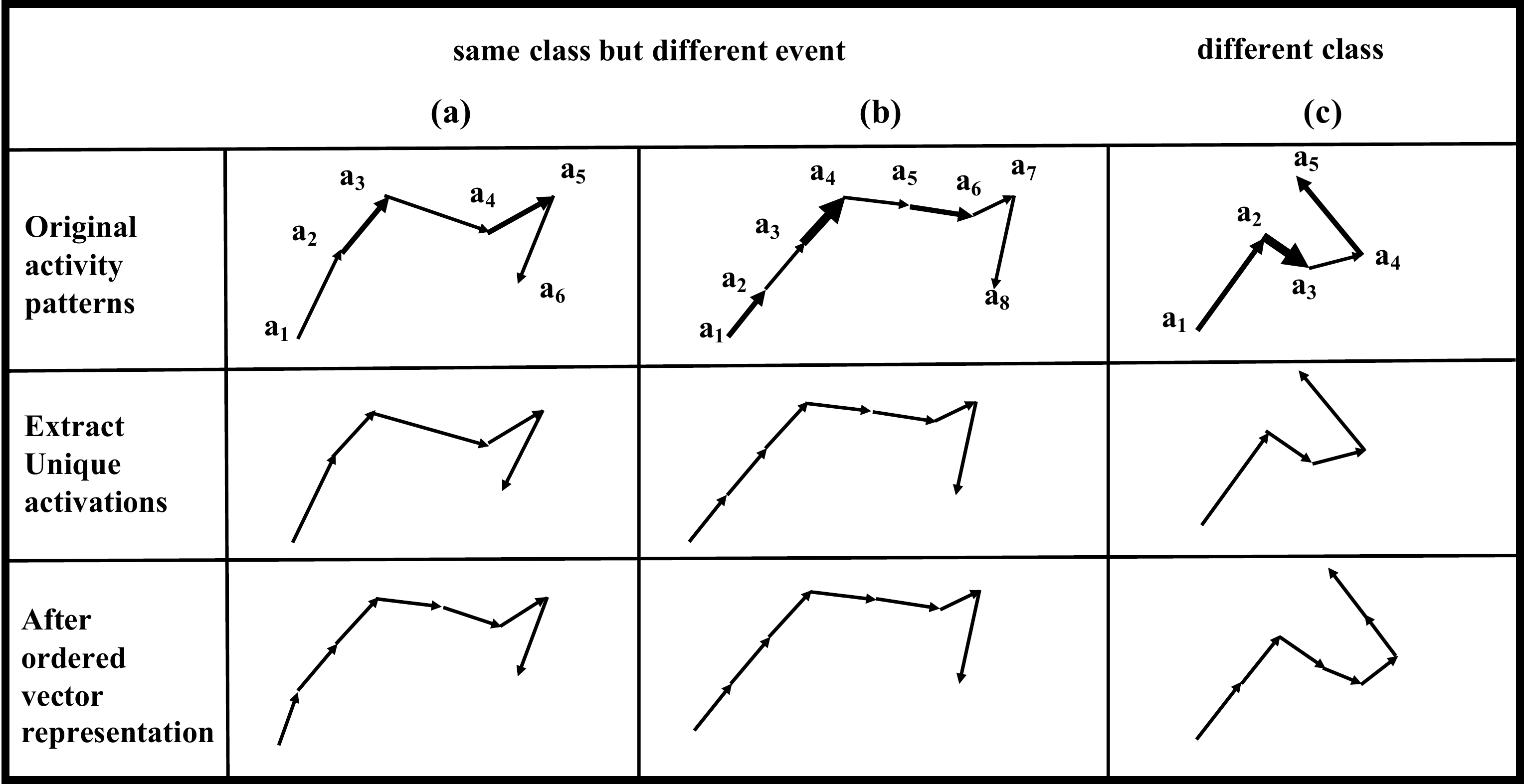}
	\caption{Application of ordered vector representation mudule to generate time-invariat pattern vectors. Top row shows the original activity patterns of the same action performed with different speed. Fast performance has less number of activated neurons than the slow one. The middle row shows the same activity patterns after extracting unique activations and the bottom row shows the pattern vectors when the number of activated neurons are becoming similar by assigning new activations.}
	\label{fig:supImp}      
\end{figure*}

Due to the nature of different actions and the speed of performing an action the activity pattern vectors have different length, which can be seen as the original activity patterns of two sequences of the same action shown in Fig.~\ref{fig:supImp}. Ordered vector representation module takes care of this feature by assigning new activations in the span of those activity pattern vectors having less number of activated neurons than the longest activity pattern vector. The new vectors will preserve the features of the original ones such as length and direction since new activations are replaced on the line connecting consecutive activations.

As illustrated by Fig.~\ref{fig:supImp}, the original activity patterns of two sequences of an action have a number of thicker arrows showing that the same neuron has been activated by similar consecutive posture frames more than once. To address this feature, first the consecutive repitition of similar activations is mapped into one unique activation and then, the activity pattern vector having the maximum number of activations is extracted and the number of its activations $K_{max}$ is calculated: 

\begin{equation} \label{eq:15}
 K_{max} = \max\limits_{v_n \in V}(k_{v_n}), 
\end{equation}

where $v_n$ is an activity pattern vector and $V$ is the set containing all activity pattern vectors. The term $k_{v_n}$ shows the number of activations of the activity pattern vector $v_n$. The goal is to increase the number of activations for all activity pattern vectors to $K_{max}$ by inserting new activations. Therefore, for each activity pattern vector, its length $l_{v_n}$ is calculated by summing the $\ell_{2}$ norm of the consecutive activations as follow:

\begin{equation}\label{eq:16}
l_{v_n}=\sum_{n=1}^{N-1} \left\Vert\left( a_{n+1}-a_{n} \rm\right) \right\Vert,
\end{equation}

where $a_n=[x_{n}, y_{n}]$ is an activation in the 2D map and $N$ is the total number of activations for the corresponding activity pattern vector. As the next step, $l_{v_n}$ is divided by the maximum number of activation $K_{max}$ from (\ref{eq:15}) to find approximately the optimal distance $delta$ between the consecutive activations of the new pattern vector:

\begin{equation}\label{eq:17}
delta = \frac{l_{v_n}}{K_{max}}.
\end{equation}

To find the location of a new insertion, the distance between a consecutive pair of activations such as $a_1$ and $a_2$ shown in Fig.~\ref{fig:supImp} (first row) is calculated as $\ell_{2}$ norm. If this value is larger than $delta$ then the new activation is inserted on the line connecting $a_1$ and $a_2$ with $delta$ distance from $a_1$. The precise location of the new insertion $a_p$ is calculated by solving the following system equations where we assume $a_1 = (x_1, y_1)$ and $a_2 = (x_2, y_2)$ and $a_p = (x_p, y_p)$.

\begin{equation}\label{eq:18}
sys = \begin{cases}
Eq_1: \frac{y_p-y_1}{y_2-y_1} = \frac{x_p-x_1}{x_2-x_1}\\
Eq_2: \left\Vert\left( a_{p}-a_{1} \rm\right) \right\Vert = delta\\
\end{cases}
\end{equation}

Having (\ref{eq:18}), $Eq_1$ shows the standard line equation in 2D space where $a_p$ must be located on and $Eq_2$ satisfies the distance criteria for the new insertion.

If the distance between $a_1$ and $a_2$ is smaller than $delta$, $a_p$ is inserted on the line connecting $a_2$ to $a_3$, which is the activation following $a_2$. Next, $a_2$ is removed from the corresponding pattern vector. This ocurs by calculating a new distance:

\begin{equation}\label{eq:19}
delta_{p} =  delta - \left\Vert\left( a_{2}-a_{1} \rm\right) \right\Vert,
\end{equation}

and solving the system equations:

\begin{equation}\label{eq:20}
sys = \begin{cases}
Eq_1: \frac{y_p-y_2}{y_3-y_2} = \frac{x_p-x_2}{x_3-x_2}\\
Eq_2: \left\Vert\left( a_{p}-a_{2} \rm\right) \right\Vert = delta_{p}\\
\end{cases}
\end{equation}

The insertion of new activations continues until the total number of activations for the corresponding activity pattern vector matches $K_{max}$.

\subsection{Labeling layer}

The output layer of the architecture is one-layer supervised neural network, which receives the activity traces of the second-layer growing grid as its input and allocates the correct action labels to the categories obtained by training the second-layer growing grid. The output layer consists of a vector of $N$ number of neurons and a fixed topology. The number $N$ is determined by the number of action categories. Each neuron $n_{i}$ is associated with a weight vector  $w_{i}\in{R}^n$. All the elements of the weight vector are initialized by real numbers randomly selected from a uniform distribution between 0 and 1, after which the weight vector is normalized, i.e. turned into unit vectors. At each learning step, a neuron $n_{i}$ receives an input vector $x(t)\in{R}^n$.

The activity $y_{i}$ in the neuron $n_{i}$ is calculated using the standard cosine metric:

\begin{equation}\label{eq:7}
y_{i}=\frac{x(t)\cdot w_{i}(t)}{||x(t)|| \, ||w_{i}||}.
\end{equation}

During the learning phase the weights $w_{i}$ are adapted by:

\begin{equation}
w_{i}(t+1)=w_{i}(t)+\beta x(t)[y_{i} - d_{i}],
\end{equation}

where the parameter  $\beta$ is the adaptation strength and $d_{i}$ is the desired activity for the neuron $n_{i}$.

\section{Experiments}
\label{sec:exp}
The categorization capacity of the growing grid architecture shown in Fig. ~\ref{fig:HGG} has been evaluated in four experiments that are described in this section. The actions used in these experiments are shown in Table \ref{tab:Table2}.

\subsection{Parameter settings}
\label{paramset}

All settings for the system hyperparameters used to train the architectures involved in performing the experiments are shown in Table \ref{tab:Table1}. Next, I will describe the approaches for setting two critical parameters of the growing grid network, the \textit{adaptation step ($\lambda$)} and the  \textit{tunning step ($\gamma$)}.

\subsubsection{Adaptation Step} In the growing grid network the \textit{adaptation step ($\lambda$)} determines the time of a new insertion. This requirs to know the total number of the input signals received by the growing grid. For better explanation, the input to the first-layer growing grid is the randomly selected action instances in which each instance is composed of a consecutive series of posture frames. for the first experiment, on average an instance of an action contains $40$ posture frames and in total there are around $10000$ posture frames for all action instances of the training set . 

In order to have a better distinction between different action samples the adaptation step $\lambda$ should be set with a value of $40< {\lambda}<10000$. The selected value of $\lambda$ in the first-layer growing grid is equal to $4300$, which is almost in the middle of the interval, the \textit{middle approach}. With this value of $\lambda$ either a complete row or a complete column is inserted almost twice in each training epoch of the growth phase (one epoch is counted when all input signals are received by the network once). Certainly the network receives all the input signals sufficiently often in the adaptation interval. 

In the second-layer growing grid of the first experiment, there are totally $217$ action pattern vectors representing the input space while there are on average $20$ input signals representing each action category. Therefore $\lambda$ should have a value in the range of $20< {\lambda}< 217$ because the aim is to better distinguish between different action categories (intra-class) than in one action category (inter-class). As a result, $\lambda$ is set to $100$ for the second-layer growing grid, which is again based on middle approach. 

\subsubsection{Tunning Step} 
\begin{figure*}
	\centering
	\includegraphics[width=0.80\textwidth]{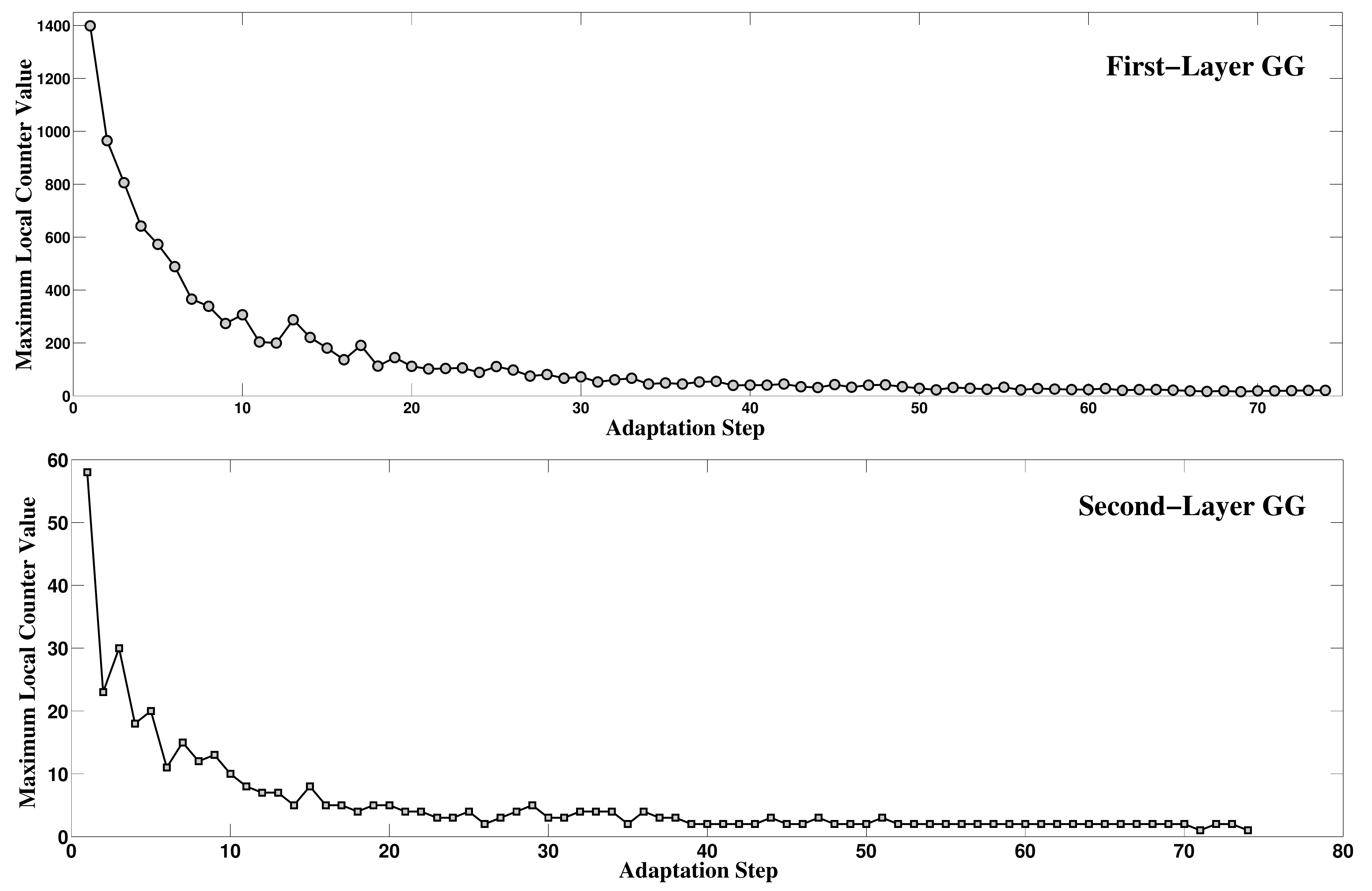}
	\caption{The maximum amount of the local counter values of all neurons calculated in each adaptation step during the growth phase of the first-layer growing grid (upper window) and second-layer growing grid (lower window).}
	\label{fig:wincount}       
\end{figure*}

Another critical parameter of the grwoing grid network is the \textit{tuning step ($\gamma$)}, which represents the performance criterion determining when the growth phase ends. One way of setting it, is through the local counter variable throughout the growth phase (see Fig.~\ref{fig:wincount}). In the beginning of the growth phase there is a maximum local counter value due to the lack of neural areas to represent the input space. By insertion of new neurons and expansion of the neural map, the local counter value decreases and finally it reaches a constant minimum value. This final state shows that there is no longer a high contrast in the activation level of different areas of the map and instead there is a homogeneous activation pattern distributed through the map. As a result of this condition the insertion of new neurons will not benefit the representational demands but it will waste the time and processing power so we have to stop it. 

Another way to determine the $\gamma$ is to set the maximum number of neurons building the neural map based on our available resources. Using this approach, the tuning step ($\gamma$) represents the maximum number of neurons. The tuning step ($\gamma$) is thus set to $900$ and $1600$ for the both first- and second-layer growing grid maps of the first experiment.

\begin {table}
\centering
\caption {Settings of the parameters of SOM and growing grid architectures used in the experiments performed in this article. $\lambda$ and $\gamma$ are the parameters defined in design and implementation of the growing grid architecture (see \ref{secgg}) and they are not applicable for the SOM architecture. Similar number of Neurons of the corresponding layers represents the same level of complexity for both architectures.} \label{tab:Table1}
\begin{tabular}{ |p{2cm}||p{2cm}|p{2cm}|p{2cm}|p{2cm}|  }
	\hline
	\multicolumn{5}{|c|}{Parameter Setting} \\
	\hline
	Parameters         & SOM.1        & SOM.2      & GG.1       & GG.2 \\
	\hline
	Neurons          & 900/2500     & 1600/2500  & 900/2500   & 1600/2500\\
	$\sigma$         & $10^{6}$       & $10^{3}$     & $10^{6}$     & $10^{3}$\\
	Soft-Max Exp     & 10          & 10         & 10         & 10\\
	Learning Rate    & $10^{-1}$      & $10^{-1}$    & $10^{-1}$    & $10^{-1}$\\
	Metric           & Euclidean    & Euclidean  & Euclidean  & Euclidean\\
	$\lambda$        & -            & -         & Middle     & Middle\\
	$\gamma$         & -            & -         & 900/2500   & 900/2500\\
	\hline
\end{tabular}
\end{table}

\subsection{ Datasets}

In this section, I will introduce the datasets used for running the experiments and some of their features. All of these datasets are available online and they provide 3D joints skeleton information.

\subsubsection{MSRAction3D Dataset} The first dataset is the MSR-Action3D dataset, \cite{MSR}. This dataset is composed of the consecutive posture frames of a human skeleton represented by 3D joint positions captured by a Kinect-like sensor. The dataset contains 563 action sequences achieved from 20 different actions performed by 10 different subjects in 2 to 3 different events. Each action sequence consists of consecutive posture frames each posture frame contains 20 joint positions expressed in 3D Cartesian coordinates. The actions of this data set are shown in Table. \ref{tab:Table2}.

\subsubsection{UTKinect Dataset}

Secondly, UTKinect data set is used \cite{xia2012view}. The videos used for this dataset are captured using a single stationary Kinect. There are $200$ action sequences consisting of 10 different action categories (shown in Table \ref{tab:Table2}). The actions are performed by 10 different subjects. Each subject performs each actions two times. Three channels were recorded: RGB, depth and skeleton joint locations. The three channel are synchronized. The frame-rate is 30f/s. The 3D skeleton data is used in this article, which contains the cartesian coordinates $[x, y, z]$ of 20 skeleton joints. The x, y, and z are the coordinates relative to the sensor array, in meters so the input to the system has 60 dimensions.

\subsubsection{Florence3DActions Dataset}

In the third experiment, the Florence3DActions dataset \cite{Seidenari} is used. This data set collected at the University of Florence during 2012, has been captured using a Kinect camera. It contains 9 different actions as shown in Table \ref{tab:Table2}. The actions are performed by 10 different subjects in 2/3 events. This resulted in a total of 215 action sequences consist of cartesian coordinates $[x, y, z]$ of 15 skeleton joints with x, y, and z coordinates relative to the sensor array. Therefore the input to the system has 45 dimensions.

\begin {table}[h!]
\centering
\caption {The action categories used in four experiments of this article.} \label{tab:Table2} 
\begin{tabular}{ | l ||  p{8cm} |}
	\hline
	\centering
	\bf{Datasets} &  \bf{Actions} \\ \hline
	\bf{MSRAction3D P.1} & 1.Hand Clap, 2.Two Hands Wave, 3.Side Boxing, 4.Forward Bend, 5.Forward Kick, 6.Side Kick, 7.Still Jogging, 8.Tennis Serve, 9.Golf Swing, 10.Pick up and Throw \\ \hline
	\bf{MSRAction3D P.2} &  1.Hand Clap, 2.Two Hands Wave, 3.Side Boxing, 4.Forward Bend, 5.Forward Kick, 6.Side Kick, 7.Still Jogging, 8.Tennis Serve, 9.Golf Swing, 10.Pick up and Throw, 11.High Arm wave, 12.Horizontal Arm Wave, 13.Using Hammer, 14.Hand Catch, 15.Forward Punch, 16.High Throw, 17.Draw X-Sign, 18. Draw Tick, 19. Draw Circle, 20.Tennis Swing\\ \hline
	\bf{UTKinnect} &  1.Walk, 2.Sit down, 3.Stand up, 4.Pick up, 5.Carry, 6.Throw, 7.Push, 8.Pull, 9.Wave Hands, 10.Clap Hands \\
	\hline
	\bf{Florence3DActions} &  1.Wave, 2.Drink from a Bottle, 3.Answer Phone, 4.Clap Hands, 5.Tight Lace, 6.Sit down, 7.Stand up, 8.Read Watch, 9.Bow  \\
	\hline
\end{tabular}
\end{table}

\subsection{Experiment 1}
\label{exp1}
In the first experiment the ability of the proposed architecture to categorize actions is tested by using the MSRAction3D dataset (\cite{MSR}).  The experiment starts with a subset of dataset containing 10 action categories performed by the whole body of the performer (arms as well as legs) so the input space represents sufficient variability by being distributed throughout the whole body of the performers. For the second part of this experiment the number of action categories are doubled to 20 different actions to test how the SOM and GG architectures will perform to categorize action samples into more classes.

\subsubsection{Part 1}
\label{sec:msr2}
The dataset used in this part, contains $287$ action samples with $10$ different actions (see Table. \ref{tab:Table2}) performed by $10$ different subjects in $2$ or $3$ repetitions. To run the experiment $25\%$ of the action samples are selected randomly for the final test experiments. The remaining samples are used to train the architectures. The neural network system was trained with randomly selected instances from the training set in two phases, the first to train the first-layer growing grid and the second to train the second-layer growing grid together with the output layer. The parameters of the architecture are set to the values shown in Table. \ref{tab:Table1}.

\begin{figure}
	\centering
	\begin{subfigure}[t]{1.0\linewidth}
		\includegraphics[width=\textwidth]{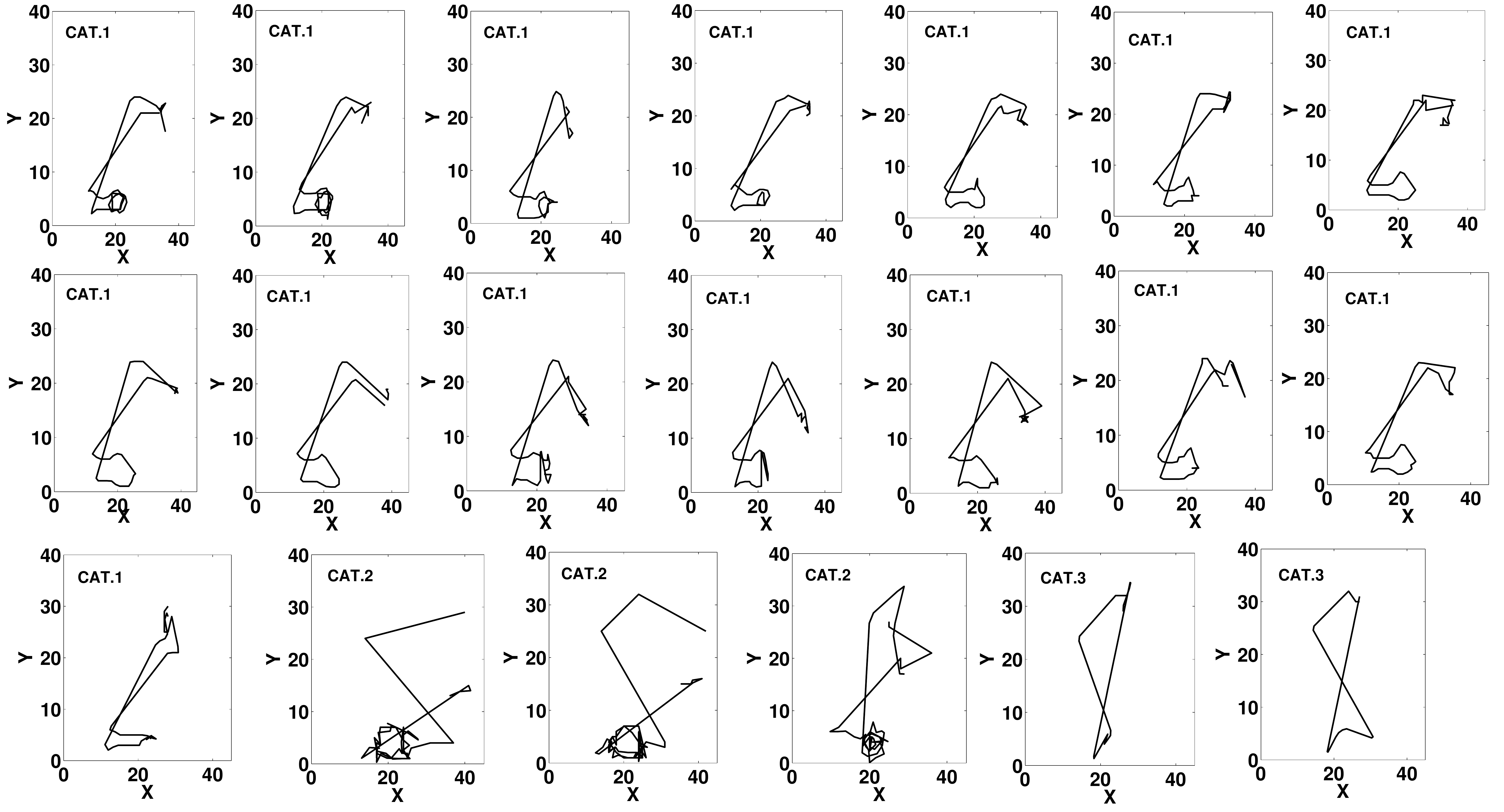}
		\caption{Train data}
		\label{fig:trpatt}
	\end{subfigure}
	\begin{subfigure}[t]{1.0\linewidth}
		\includegraphics[width=\textwidth]{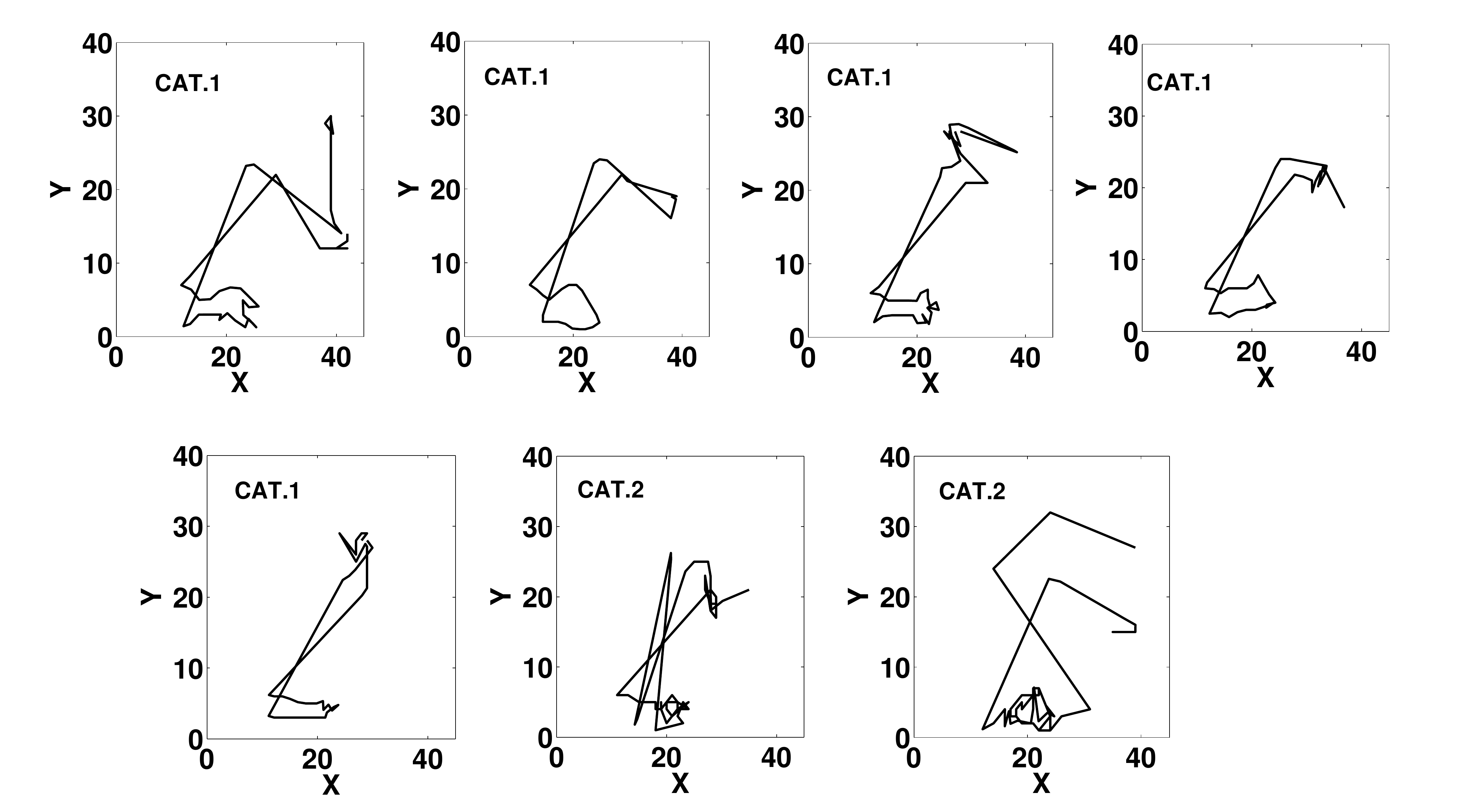}
		\caption{Test data}	
		\label{fig:tspatt}
	\end{subfigure}
	\caption{Time-invariant action patterns belonging to different sequences of action Two Hands Wave in the training dataset (a) and test dataset (b). Each block in the figure represents an action pattern of the corresponding action sequence. It can be seen that the action patterns of different sequences of the same action are similar and that the representions are plausible. The differences between the patterns show the formation of the sub-categories inside the action category, which is the result of the fact that the same action is performed in different ways by the actors. This increases the complexity of the input space and makes the categorization problem more challenging. A significant similarity can be detected between the action patterns in (b) and the ones represented in (a).}
	\label{fig:patterns}
\end{figure}

By applying the input data to the trained growing grid of the first-layer, the elicited activity traces of the actions are extracted and applied to the ordered vector representation layer. The purpose is to generate the corresponding time-invariant action patterns (see \cite{Gharaee5}). These patterns are illustrated in Fig.~\ref{fig:patterns} for the training data (a) and for the test data (b) of the action Two Hands Wave. The similarity of the patterns corresponding to the action Two Hands Wave strengthens the fact that the growing grid network represents the input space in an efficient way by extracting the most salient features of the input data. The patterns of the test dataset, which the trained network has never seen before, represents the generalizability and robustness of the system.

\begin{figure}
	\centering
	\begin{subfigure}[t]{1.0\linewidth}
		\includegraphics[width=\textwidth]{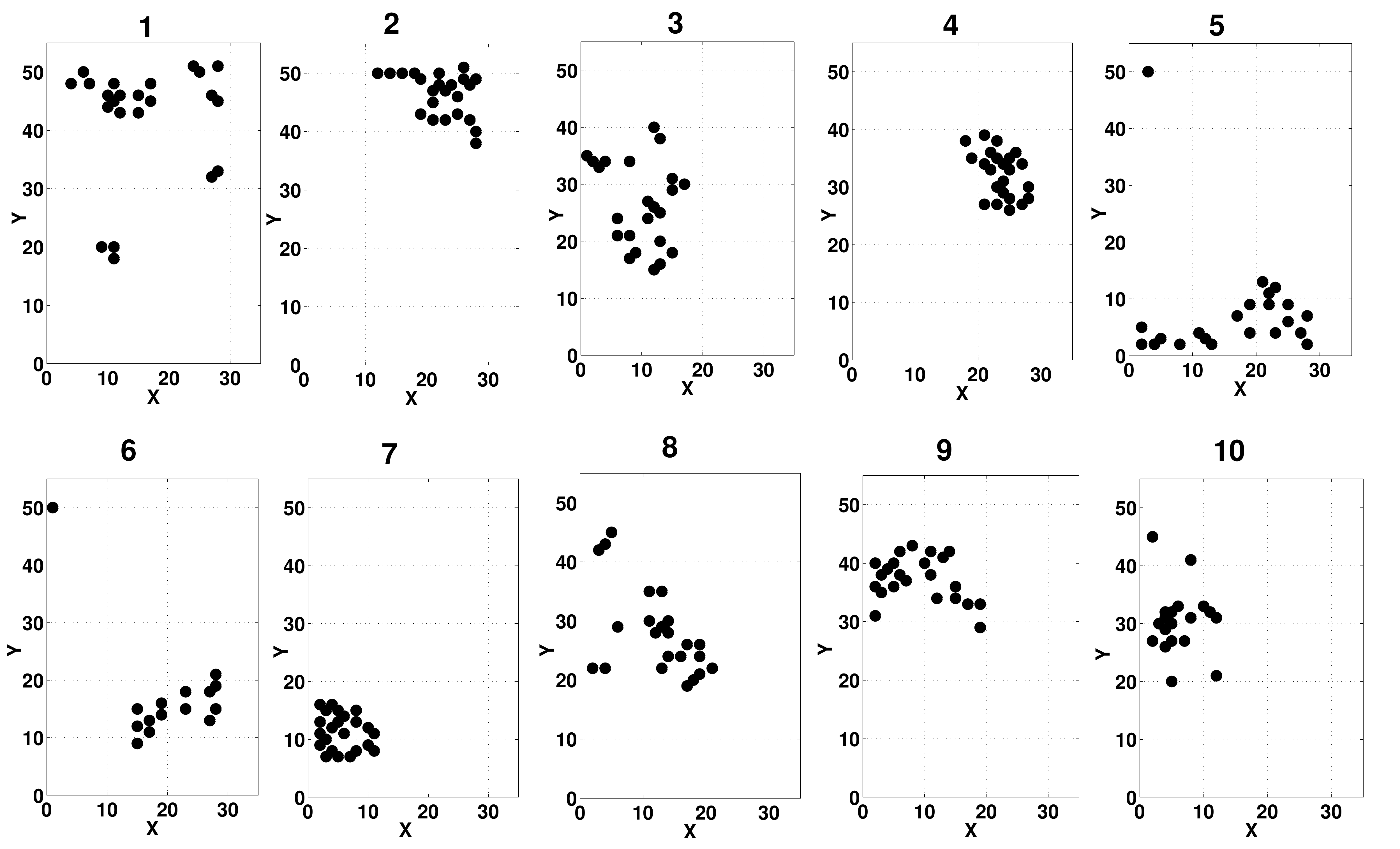}
		\caption{Train data}
		\label{fig:trclass}
	\end{subfigure}
	\begin{subfigure}[t]{1.0\linewidth}
		\includegraphics[width=\textwidth]{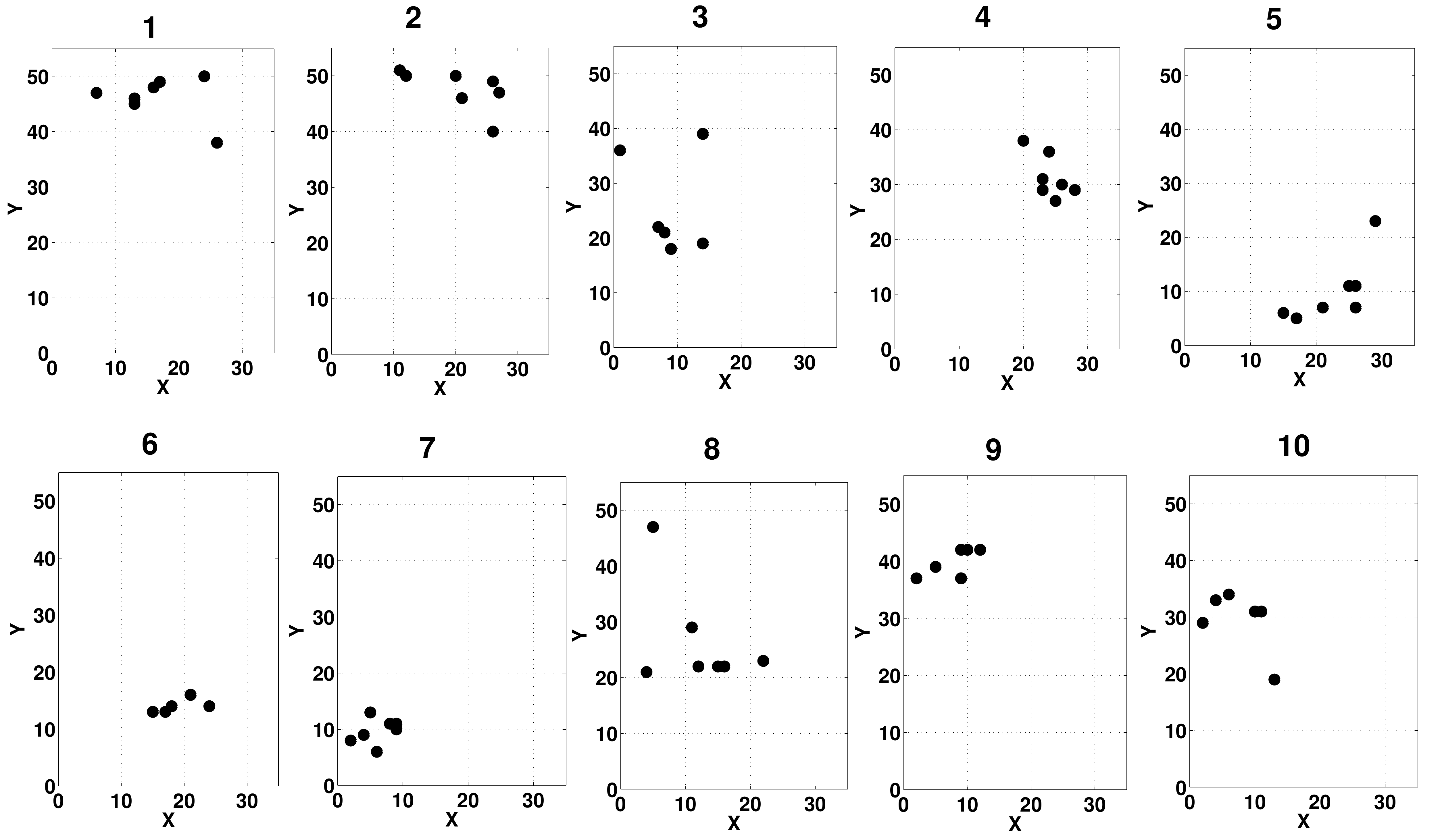}
		\caption{Test data}	
		\label{fig:tsclass}
	\end{subfigure}
	\caption{The clustering of the second-layer growing grid when it receives action pattern vectors of train data (a) and test data (b) as its input space. Each block of the figure shows the formed category of a particular action. The actions are: 1. Hand Clap, 2. Two Hands Wave, 3. Side Boxing, 4. Forward Bend, 5. Forward Kick, 6. Side Kick, 7. Still Jogging, 8. Tennis Serve, 9. Golf Swing, 10. Pick Up and Throw. It can be seen that different areas in the map represent different action categories. It is also shown that the network allocates the same areas of the map shown in (a) to the corresponding action categories of the test data shown in (b).}
	\label{fig:class}
\end{figure}

By training the second-layer growing grid on the action patterns, the second part of the architecture is designed to categorize the actions. Fig.~\ref{fig:class} shows the clusters of all $10$ different actions that have been created in the trained growing grid for the training data in Fig.~\ref{fig:class} (a) and for the test data in Fig.~\ref{fig:class} (b). As shown in these figures, the action categories are separated and almost different areas of the map are allocated to different categories.

To better show how the system recognizes different action samples, a confusion matrix for the performance result of this experiment is shown by Fig.\ref{fig:configs1} (a). Based on the confusion matrix, mis-classification occurs for a few samples of the actions \textit{Golf Swing}, \textit{Hand Clap}, \textit{Tennis Serve} and \textit{Pick up and Throw}, while in almost other cases the system recognizes the correct action.

\subsubsection{Part 2}
For the second experiment, the entire MSR Action 3D dataset \cite{MSR} is used as the input. This set contains $20$ actions as shown in Table.\ref{tab:Table2} and in total $563$ action instances. To run this experiment $25\%$ of the dataset is selected randomly for the test experiments and the remaining is used to train the architectures.

To train the growing grid architecture shown in Fig.~\ref{fig:HGG}, all the parameters are set based on the values shown in the Table. \ref{tab:Table1}. The number of neurons of growing grid networks is set to $2500$ for this experiments. The double number of action classes require a larger map to represent the input space. Therefore, the tuning step ($\gamma$) is also set to $2500$ for this experiment.

\begin{figure}
	\centering
	\begin{subfigure}[t]{0.43\linewidth}
		\includegraphics[width=\textwidth]{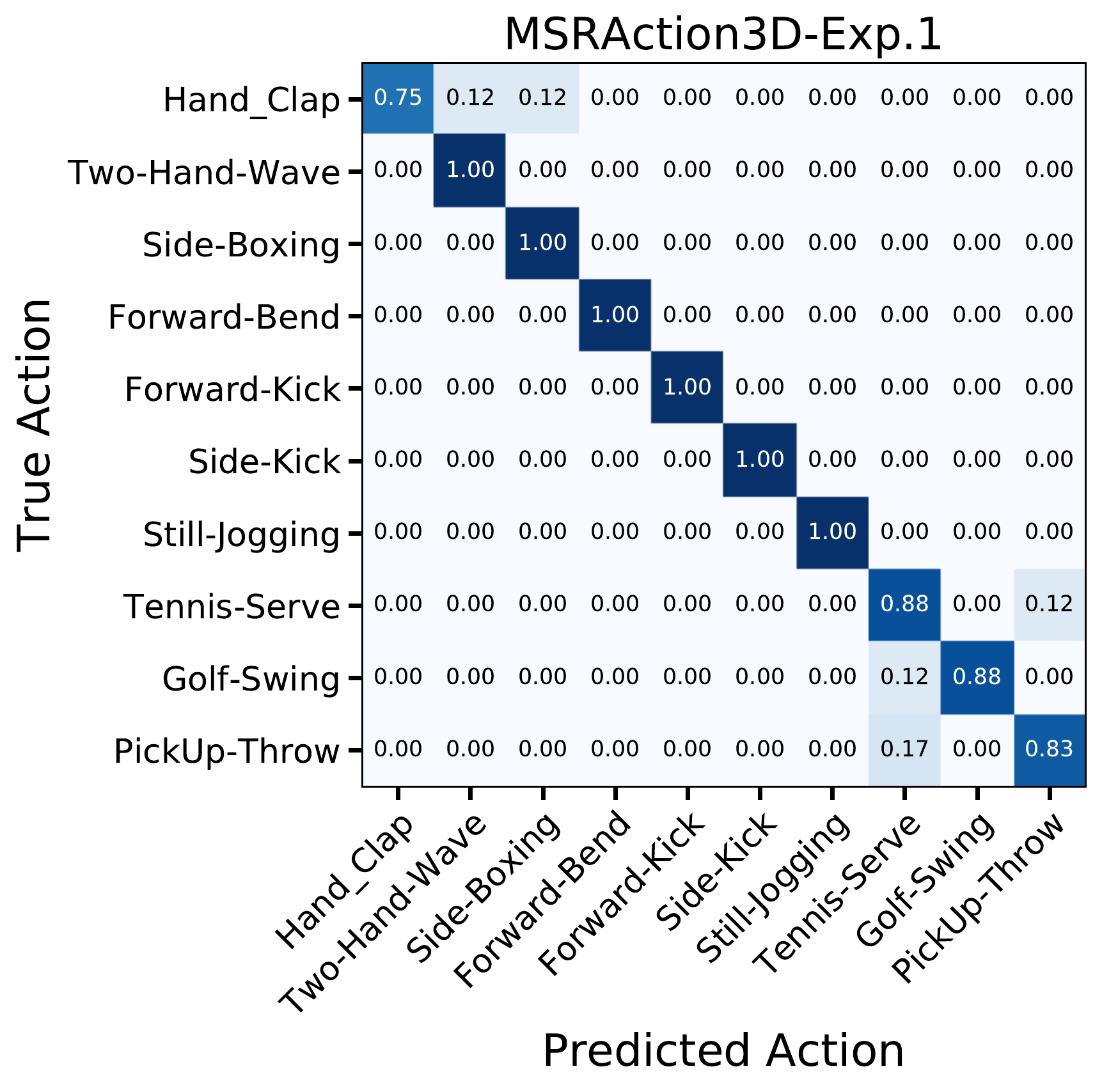}
		\caption{}
		\label{fig:config1}
	\end{subfigure}
	\begin{subfigure}[t]{0.56\linewidth}
		\includegraphics[width=\textwidth]{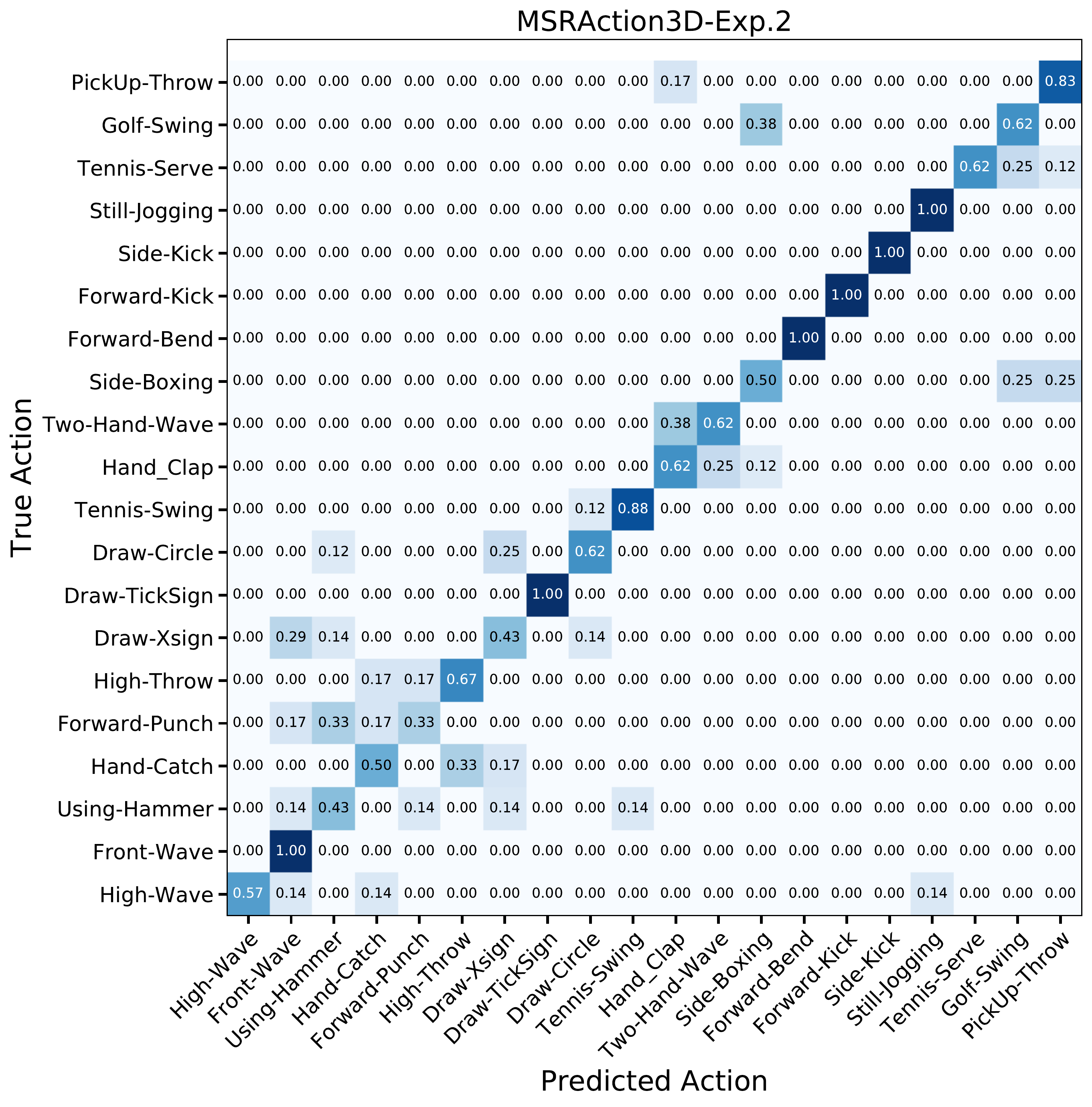}
		\caption{}	
		\label{fig:config2}
	\end{subfigure}
	\caption{The Confusion Matrix showing the results of Experiment 1 using MSRAction3D dataset as the input. The classification test results of the 10 actions performed in the first part of Experiment 1 (a) and the classification test results of the 20 actions performed in the second part of Experiment 1 (b).}
	\label{fig:configs1}
\end{figure}

For better illusteration of the classification results, the confusion matrix for the test
data is shown in Fig.~\ref{fig:configs1}. As shown by Fig.~\ref{fig:configs1} (b), the mis-classification occurs more often in the actions performed using upper part of the body such as \textit{High Wave}, \textit{Front Wave}, \textit{Using Hammer} and \textit{Hand Catch}. The reason exists in the intra-class similarity while performing the actions using the same body parts such as the arms. This results in more action samples with similar components represented by the posture frames compare to when the actions perform using the whole body, arms as well as the legs.

\subsection{Experiment 2}
\begin{figure}
	\centering
	\begin{subfigure}[t]{0.48\linewidth}
		\includegraphics[width=\textwidth]{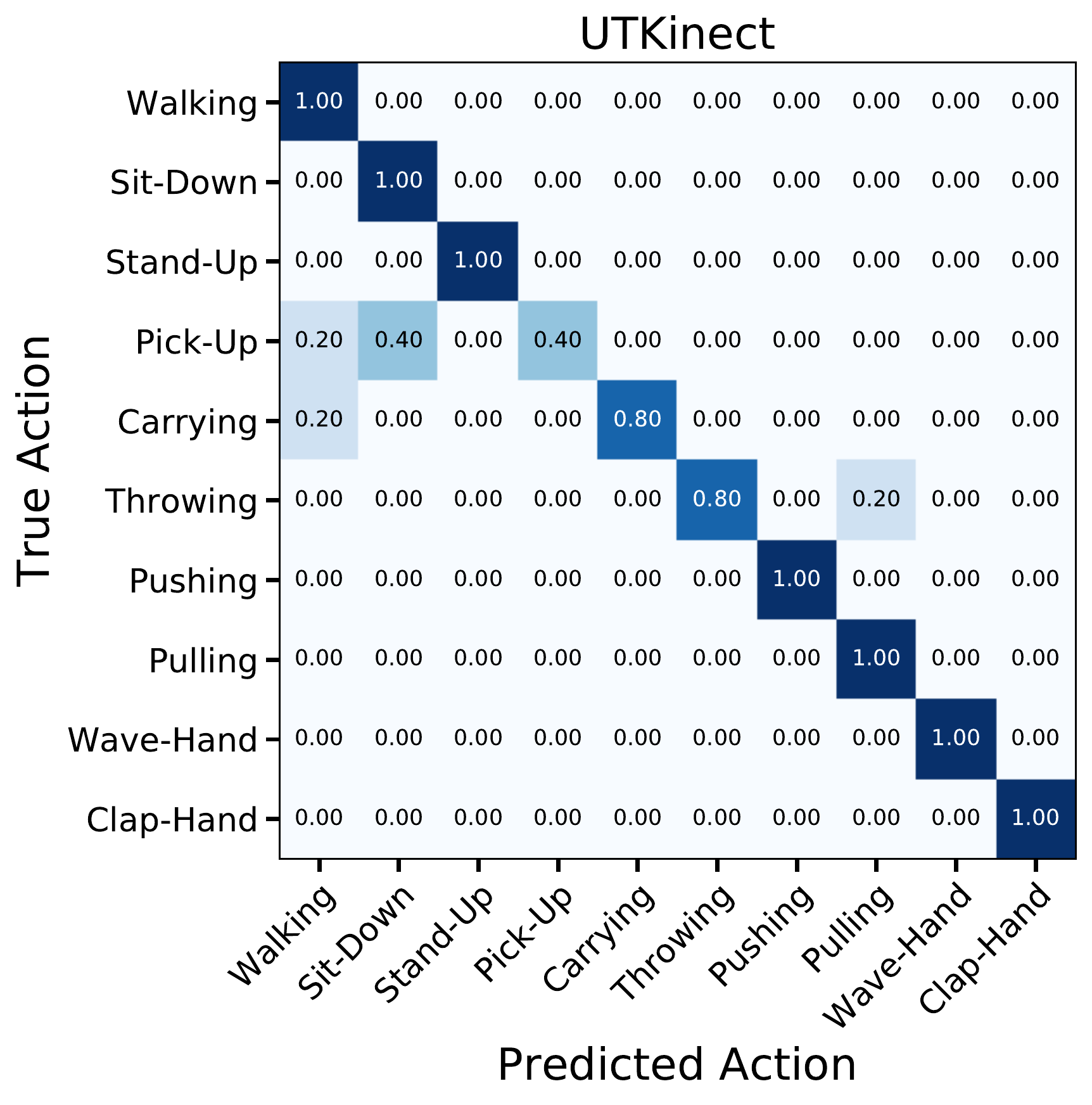}
		\caption{}
		\label{fig:config3}
	\end{subfigure}
	\begin{subfigure}[t]{0.48\linewidth}
		\includegraphics[width=\textwidth]{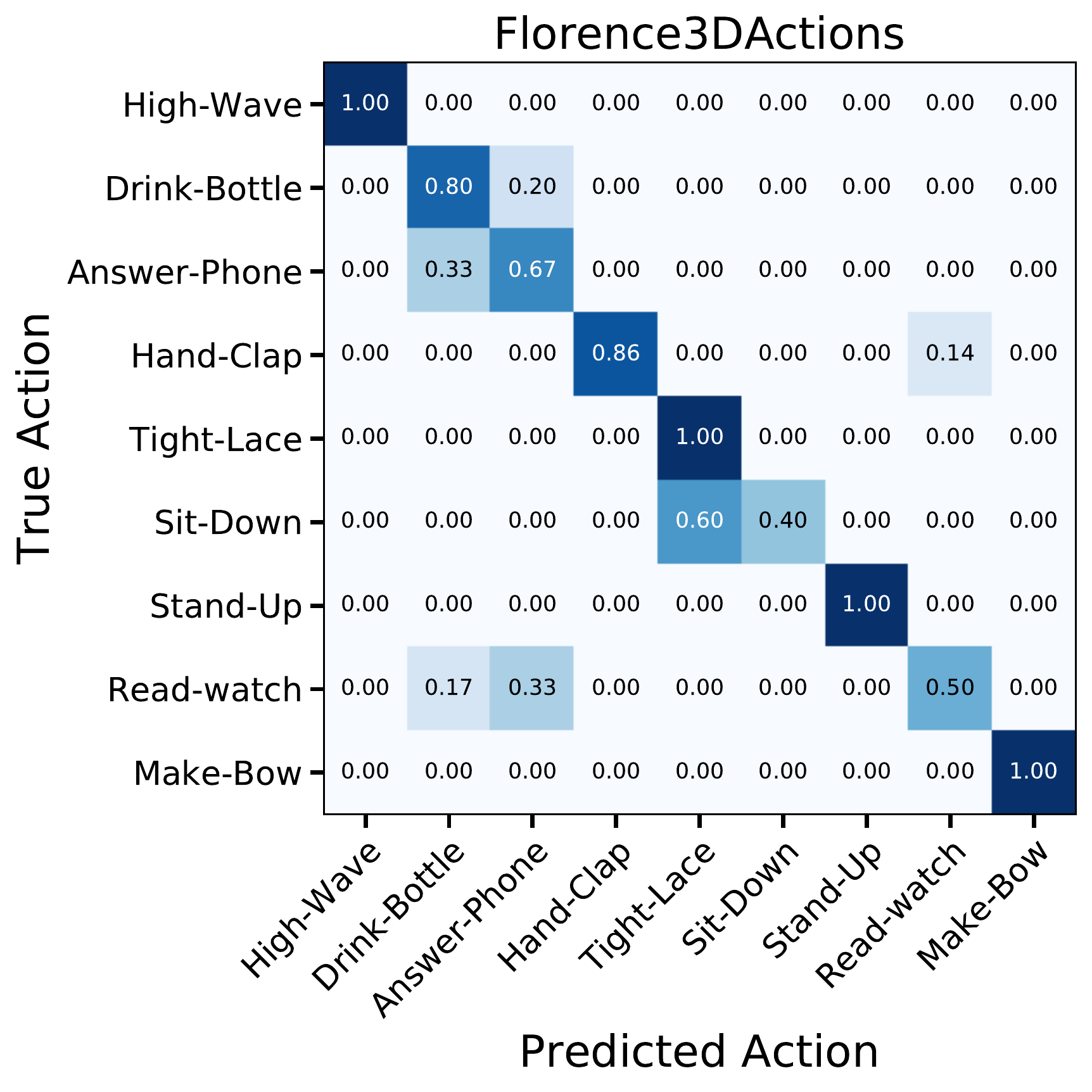}
		\caption{}	
		\label{fig:config4}
	\end{subfigure}
	\caption{The Confusion Matrix showing the results of Experiment 
		2 and Experiment 3 using UTKinect and Florence3DActions datasets as the inputs. The classification test results of the 10 actions performed in the Experiment 2 (a) and the classification test results of the 9 actions performed in the Experiment 3 (b).}
	\label{fig:configs2}
\end{figure}

For this experiment the UTKinect dataset is used \cite{xia2012view} with $200$ action samples from $10$ different action categories shown in Table.\ref{tab:Table2}. The system parameters are set similar to the first experiments as shown in Table \ref{tab:Table1}. To run this experiment, 10-fold cross validation method is used. The confusion matrix illusterating the classification results of the test data of this experiment is shown in Fig.~\ref{fig:configs2} (a). The mis-classification occurs for some samples of only three actions \textit{Pick up}, \textit{Carrying} and \textit{Throwing}. The action  \textit{Pick up} is performed by taking an object from the ground, which resembles the body postures when sitting on a chair and this could be the reason of its mis-classification with the action \textit{Sit Down}.

\subsection{Experiment 3}
In the third experiment, the Florence3DActions dataset \cite{Seidenari} is used as the input with $215$ action samples from $9$ different action categories shown in Table.\ref{tab:Table2}. For running this experiment, the system parameters are set similar to the values shown in Table \ref{tab:Table1} and the system is trained using 10-fold cross validation. A better illusteration of the classification test results of this experiment is represented by confusion matrix available in Fig.~\ref{fig:configs2} (b).

The mis-classification occurs more in the action samples of this experiment. There are five actions having mis-classification results. The three actions \textit{Drink Bottle}, \textit{Answer Phone} and \textit{Read Watch} confused the system mainly because in all of them the main component is to lift the arm up and if the system doesn't receive any data from the object involves in performing the action such as bottle, cell-phone and watch it only sees the same body posture and it is difficult to distinguish them. This is one of the main challanges in the action recognition task when the objects involved in performing the actions play a key part in recognizing them. The concept of manner vs result actions is studied in \cite{Gharaee4}

\section{Comparison}

To make a better evaluation of the growing grid architecture, its performance is compared with another architecture based on the self-organizing maps (SOM) developed in \cite{Gharaee5} in terms of both recognition accuracy and the learning speed. Therefore, three more experiments using MSRAction3D (the whole set), Florence3DActions and UTKinect datasets are performed. To train the SOM architecture (\cite{Gharaee5}), the parameters are set based on the values shown in the Table. \ref{tab:Table1}. The performance of SOM architecture using the MSRAction3D dataset of the experiment in section \ref{sec:msr2} is based on the \cite{Gharaee5}, while the experiments using this dataset is repeated for this article.

 The recognition result per action for the experiment using MSRAction3D (the whole set) is illustrated in Table. \ref{tab:Table3}, the accuracy of categorizing different actions while using growing grid architecture is significantly superior in almost all actions. For some actions SOM architecture performs slightly better, but the overall performance of the training data and the generalization test data for the growing grid architecture outperforms the SOM architecture.

The overall performances of the two architectures in categorizing the action sequences of different datasets are also compared in Table \ref{tab:Table4}. The accuracy of the system's capacity to categorize actions for the generalization test data, the total number of learning epochs together with the relative time of running both architectures are shown in the Table \ref{tab:Table4}. As the results show although both of the architectures use the same level of space complexity represented by the number of neurons allocated to first and second layer SOM/GG, the growing grid architecture outperforms the SOM architecture both in accuracy as well as learning speed. The improvements of the learning speed is quite better than the accuracy while the system learns 3 to 4 times faster when using growing grid instead of SOM. 

A reason for the improvement of the learning speed is the prior knowledge of the input space that the growing grid networks gain during the growth phase. Moreover the growth phase starts with a small number of neurons and the grid grows gradually during adaptation intervals and as a result the learning speeds up. The learning occurs faster in the beginning iterations due to the smaller size of the map. In contrast, in the SOM implementation there is a preset size of the network and the system processes the information in the whole learning phase with this fixed size. As a consequence, the efficiency of the SOM decreases in realistic problems with more complicated and diverse input spaces, which require larger grids of neurons (see also \cite{Fritzke4}).

The recognition accuracy of the proposed growing grid architecture that has been achieved in this experiment is among the highest compared to the methods tested on the same or even a smaller number of action categories (see \cite{Du}, \cite{Veeriah}, \cite{Chaudhry, li2017mining, Oreifej, Xia1, Yang, Wang2, Wang-Jiang}).

\begin {table}
\centering
\caption {Results of the two architectures shown in the numbers representing the recognition accuracy of each actions in the train and test data. The last row of the table shows the average accuracy in different experimental conditions.} \label{tab:Table3} 

\begin{tabular}{l*{5}{c}}
\centering
\bf Actions              & \bf Train-SOM & \bf Train-GG &  \bf Test-SOM & \bf Test-GG \\
\hline
High Arm Wave                     & 92.00\% & 99.30\% & 42.80\% & 57.10\%   \\
Horizontal Arm Wave            & 85.00\% & 100\% & 35.80\% & 100\%   \\
Using Hammer                     & 88.00\%  & 100\% & 28.60\% & 42.90\%   \\
Hand Catch                         & 83.20\%  & 100\% & 50.00\% & 50.00\%   \\
Forward Punch                   & 82.10\% & 99.20\% & 28.60\% & 33.20\%   \\
High Throw                       & 90.50\% & 100\% & 46.40\% & 66.80\%   \\
Draw X Sign                     & 89.00\% & 100\% & 39.30\% & 43.10\%   \\
Draw Tick                        & 92.20\% & 100\% & 71.40\% & 100\%   \\
Draw Circle                     & 95.70\% & 100\% & 50.00\% & 61.90\%   \\
Tennis Swing                  & 97.40\% & 100\% & 35.80\% & 88.00\%   \\
Hand Clap                      & 95.70\% & 98.60\% & 67.80\% & 62.00\%   \\
Two Hands Wave           & 98.30\% & 99.30\% & 82.00\% & 62.00\%   \\
Side Boxing                   &98.30\% & 100\% & 42.90\% & 50.00\%   \\
Forward Bend               & 100\% & 100\% & 100\% & 100\%   \\
Forward Kick                & 100\% & 100\% & 96.40\% & 100\%   \\
Side Kick                      & 96.00\% & 100\% & 96.40\% & 100\%   \\
Still Jogging                & 100\% & 100\% & 100\% & 100\%   \\
Tennis Serve              & 98.30\% & 100\% & 82.00\% & 62.00\%   \\
Golf Swing                    & 97.40\% & 100\% & 57.50\% & 62.00\%   \\
Pick Up and Throw       & 94.00\% & 100\% & 39.30\% & 83.00\%   \\
\bf Total Average       & \bf 94.00\% & \bf 99.80\% & \bf 59.61\% & \bf 71.20\%   \\
\hline
\end {tabular}
\end {table}

\begin {table}
\centering
\caption {Comparing the performance of SOM and growing grid architectures. Acc denotes the recognition accuracy of the generalization test data. The Ep shows Epoch, which is the total number of times all input signals have been received by the system to train its parameters (one epoch is counted when all input signals are received by the network once). RT is the relative time and it shows the proportional time duration required to train the architecture.} \label{tab:Table4} 
\begin{tabular}{  |c| c c c | c c c |}
	\cline{1-7}
	\centering
	\multirow{3}{7em}{\bf{Dataset}}& \multicolumn{3}{|c|}{\bf{SOM}}&\multicolumn{3}{|c|}{\bf{Growing Grid}} \\  
	\cline{2-7}
	&  Acc & Ep  & RT  & Acc & Ep & RT \\
	\hline
	\multicolumn{1}{ |c|  }{MSRAction3D (1) \cite{MSR}}  & 90.00\%    & 1300 & 0.83 & 93.00\%  & 200  & 0.17\\
	\hline
	\multicolumn{1}{ |c|  }{MSRAction3D (2) \cite{MSR}}  & 59.61\% & 1600 & 0.81 & 71.20\% & 250  & 0.19\\
	\hline
	\multicolumn{1}{ |c|  }{UTKinect \cite{xia2012view}}          & 87.31\% & 1600 & 0.84 & 90.00\% & 300  & 0.15\\
	\hline
	\multicolumn{1}{ |c|  }{Florence3DActions \cite{Seidenari}} & 75.50\%  & 1600 & 0.77 & 80.10\%  & 300  & 0.23\\
	\hline
\end{tabular}
\end{table}

\section{Discussions}

In this section we elaborate the advantages and disadvantages of the proposed growing grid architecture. If we start with modeling action perception with the aid of teleological representations, that is, the expression of the cause and effect of the action performed, as proposed in \cite{Lallee}, the comprehension of actions that have an effect on an independent entity in the world is improved. In many other actions, such as \textit{point}, \textit{wave} and \textit{walk}, there is no effect visible to the observer. For these actions the spatial trajectory of the performer is the only resource of information. The action recognition system proposed in this paper receives the spatial trajectory of the performer as the input and makes the analysis of what action is performed based on this information so it only relies on the performer for recognizing the action. 

The still image technique for action recognition proposed in \cite{WeilongYang} depends to a large extent on the quality of the images. Since less information is available in the 2D still images, compared to the 3D joint postures utilized in the current study, the system's performance is influenced. Moreover there are actions such as \textit{Draw X-Sign} that can only be identified by using a series of consecutive key postures, which is not possible with the aid of one still image. Most actions have spatio-temporal characteristics and require a system that receives consecutive postures in a temporal order to make a correct recognition. The growing grid architecture meets this requirement by extracting the spatio-temporal features of the actions from the consecutive posture frames. 

The growing grid action recognition architecture resembles the learning mechanisms of the human cortex that performs similar recognition tasks by exploiting properties such as topological representation, layered organization and synaptic plasticity. It also employs cognitive functions such as attention and dynamic extraction to complete the recognition task. This makes the system presented here more biologically realistic than other action recognition methods such as \cite{Seo}, \cite{Ballan}, \cite{Wang-Chunyu} and \cite{Li}.

In contrast to GWR method presented in \cite{parisi2015self}, the growing grid approach proposed in this article does not add or remove edges connecting the nodes of the lattice. It instead creates and expands a 2D lattice of nodes having fixed length edges connecting the neighboring nodes. The growing grid network proposed in this article also expands by adding a complete row/collumn of nodes rather than adding/removing a single node or edge when it is required as proposed by \cite{parisi2015self}. On the other hand, the system proposed in this article uses raw 3D skeleton joints as the input. While the input features fed to the GWR represent a pose-motion vector containing the two slopes angles of the upper and the lower body orientations together with the horizontal and vertical speeds. However, the experiments evaluating the performance of GWR \cite{parisi2015self} on different datasets from the ones used in this article shows the clustering precesion of $91.9\%$, which could be compared with the performance results presented in Table. \ref{tab:Table4}.

The methods proposed in \cite{Hou} and \cite{Veeriah} are neural network based approaches (CNN-based and RNN-based approaches) that utilize skeleton information as the input data. The main step in running experiments with the CNN-based approach is to convert the posture frames of skeleton data into images in which the spatio-temporal information is collected from the color and the texture of the image. While this process may decrease the quality of the data, it also adds another preprocessing level to the system.  

A main challenge of deep learning approaches (such as \cite{Hou}, \cite{Du}, \cite{Veeriah}, \cite{Ijjina}, \cite{Pigou} and \cite{Wang-Pichao}) is to encode the temporal information. Although they use temporal fusion, they tend to neglect the temporal order of the input data of actions. As a clarification, the 3D filters and the 3D pooling filters have a very rigid temporal structure and they only receive a predefined small number of frames as their input. Furthermore the optical flow methods are computationally expensive, and the methods using the sequence to images technique inevitably lose temporal information during encoding. In a growing grid architecture, the temporal information is represented with the aid of the vector patterns created in the first-layer growing grid, so the system models spatial information as well as temporal information. The spatio-temporal information is represented in the action pattern vectors, where the shapes, length and directions of the pattern vectors inform us about different spatio-temporal characteristics of the actions.

The approach proposed in this article receives 3D skeleton joints as the input space, however, the method proposed by \cite{buonamente2016hierarchies} uses black and white contours representing action sequences as the input to a hierarchies of self-organizing maps for action recognition. In another approach proposed by \cite{parisi2017lifelong}, the grayscaled images are used as the input to the system for incrementally learning to classify human actions. The approach proposed in this article could be further expanded using higher-dimensional raw input images through application of convolutional self organizing neural networks, however, this idea has not been tested in experiments yet.

\section{Conclusion}
This article has presented a novel hierarchical architecture for recognizing and categorizing human actions. The architecture utilizes two layers of growing grid neural networks and one layer of supervised neural network. The first-layer growing grid represents the input space that receives the consecutive 3D posture frames of the actions performed. In the second-layer growing grid, the action pattern vectors representing the motions are categorized. The third layer of the system labels the action categories developed by the second layer.

The pre-processing layer executes functions such as ego-centered coordinate transformation, scaling, attention focusing and dynamic extraction on the peripheral input data before it is sent to the first-layer growing grid. The ordered vector representation layer connecting the two layers of growing grid constructs time-invariant action patterns that are extracted by connecting elicited activities of the first-layer growing grid.

The performance of the SOM architecture (see also \cite{Gharaee2}, \cite{Gharaee3}, \cite{Gharaee4} and \cite{Gharaee5}) and of the growing grid system has been compared as a part of this study. The comparison results show that the performance of the growing grid system for the same task is significantly better, both in terms of accuracy and in terms of learning speed. This improvement is considerable as regards the learning speed, which is due to the prior knowledge the growing grid system gains during the growth phase.

By using the growing grid, one not only preserves the beneficial features of the self-organizing maps such as construction of layered and topographical organization maps, lateral interaction, unsupervised learning and the ability to represent high-dimensional sensory input in low-dimensional sensory maps, but also benefits from an automatic size-setting mechanism, which gives the system more flexibility and robustness and makes it learn more efficiently.

Moreover, by utilizing growing grids, the system automatically gains prior knowledge of the input space during the growth phase. By applying this information it will expand the map by inserting a new row or column wherever there is a high representational demand. This is compatible with the findings that the cortex preferentially allocates expanded areas to represent the parts of an input space that are proportionally most used (\cite{Buonomano}). In the growing grid model implemented for this study, the local counter value of each neuron determines its amount of activation accumulated during the adaptation interval and the neuron having the maximum local counter value determines the area of the map that is mostly used to represent the input space.

Handling various types of input data helps in investigating the abilities of the action recognition system in various contexts. Nevertheless, there are big challenges in running the action recognition experiments. Among them is the intra-class and inter-class similarity, which means that individuals perform the same action with different characteristics so two different actions can be distinguished only by very subtle spatio-temporal details. Moreover, when the data involves a large number of action categories, the same action can be interpreted differently in different scene contexts (for example, \textit{drink} and \textit{eat}). Finally, disturbances, such as occlusions, cluttered backgrounds, changes of illumination and viewpoints, can influence the action perception. By using the pre-processing layer, we tried to overcome some of these challenges and improve the performance of the system.

The growing grid architecture described in this paper functions on a wide range of datasets. In the two experiments, we used two different input datasets of different length and number of actions. This is another advantage of the proposed architecture compared to the  available deep learning approaches which rely on large sets of labeled training data. Providing the large labeled input data is very costly, however. In some particular cases, for example medical applications, it is impossible to build a large labeled dataset of actions.

Among the plans for the future is to develop an online version of the growing grid architecture that recognizes more varying types of actions in a real-time mode. Another plan is to implement the attention mechanism so that the attentional process can be automatically based on the salience map of the skeletal data from the human actor. By extending the input data to a wider range of actions, especially actions containing interactions between more than one performer as well as the activities that occur in a long time interval with more variations of skeletal movements, I will develop a more robust and generalized action recognition system.     

\section*{Acknowledgment}
This work was partially supported by the Wallenberg AI, Autonomous Systems and Software Program (WASP) funded by the Knut and Alice Wallenberg Foundation.

\bibliography{References}


\end{document}